\documentclass[lettersize,journal]{IEEEtran}
\usepackage{amsmath,amsfonts}
\usepackage{array}
\usepackage{textcomp}
\usepackage{stfloats}
\usepackage{url}
\usepackage{verbatim}
\usepackage{graphicx}
\usepackage{cite}

\usepackage{epstopdf}
\usepackage{subfigure}
\usepackage{multirow}
\usepackage[linesnumbered, ruled]{algorithm2e}

\usepackage{amsthm}
\usepackage{amssymb}
\usepackage{algorithmic}
\usepackage{color}
\usepackage[table]{colortbl}

\usepackage{array}
\usepackage{booktabs}
\usepackage{amsmath}

\usepackage{times}
\usepackage{tabularx}
\usepackage{array}
\usepackage{booktabs}
\usepackage{amstext}
\usepackage{relsize}
\usepackage{lipsum}
\usepackage{array}
\usepackage{slashbox}
\newtheorem{myDef}{Definition}
\newtheorem{myThe}{Theorem}

\begin{document}

\title{Asymmetric Transfer Hashing with Adaptive Bipartite Graph Learning}

\author{Jianglin~Lu, Jie~Zhou,~\IEEEmembership{Member,~IEEE,}, Yudong~Chen, Witold~Pedrycz,~\IEEEmembership{Life Fellow,~IEEE}, Kwok-Wai~Hung  
\thanks{Manuscript received March 14, 2022. This work was supported by the National Natural Science Foundation of China (No. 62076164),  Guangdong Basic and Applied Basic Research Foundation (No. 2021A1515011861), Shenzhen Science and Technology Program (No. JCYJ20210324094601005), and Shenzhen University Overseas Study Fund (named after the endowment of the Talent Fund). (\textit{Corresponding author: Jie Zhou}) }
\thanks{Jianglin Lu is with the College of Computer Science and Software Engineering, Shenzhen University, Shenzhen, Guangdong 518060, China and also with the Department of Electrical and Computer Engineering, Northeastern University, Boston, MA 02115 USA (email: jianglinlu@outlook.com).}
\thanks{Jie Zhou is with the College of Computer Science and Software Engineering, Shenzhen University, and SZU Branch, Shenzhen Institute of Artificial Intelligence and Robotics for Society, Shenzhen, Guangdong 518060, China (email: jie\_jpu@163.com).}
\thanks{Yudong Chen is with the School of Information Technology and Electrical Engineering, University of Queensland, Brisbane, QLD 4072, Australia (e-mail: yudong.chen@uq.edu.au).}
\thanks{Witold Pedrycz is with the Department of Electrical \& Computer Engineering, University of Alberta, Edmonton, Canada, and the Systems Research Institute, Polish Academy of Sciences, Warsaw, Poland (e-mail: wpedrycz@ualberta.ca).}
\thanks{Kwok-Wai~Hung is with Tencent Music Entertainment, Shenzhen, Guangdong 518060, China (e-mail: guoweihung@tencent.com).}
}

\markboth{Journal of \LaTeX\ Class Files,~Vol.~14, No.~8, August~2021}%
{Shell \MakeLowercase{\textit{et al.}}: A Sample Article Using IEEEtran.cls for IEEE Journals}


\maketitle

\begin{abstract}
Thanks to the efficient retrieval speed and low storage consumption, learning to hash has been widely used in visual retrieval tasks. However, the known hashing methods assume that the query and retrieval samples lie in homogeneous feature space within the same domain. As a result, they cannot be directly applied to heterogeneous cross-domain retrieval. In this paper, we propose a \textit{Generalized Image Transfer Retrieval (GITR)} problem, which encounters two crucial bottlenecks: 1) the query and retrieval samples may come from different domains, leading to an inevitable {domain distribution gap}; 2) the features of the two domains may be heterogeneous or misaligned, bringing up an additional {feature gap}. To address the GITR problem, we propose an \textit{Asymmetric Transfer Hashing (ATH)} framework with its unsupervised/semi-supervised/supervised realizations. Specifically, ATH characterizes the domain distribution gap by the discrepancy between two asymmetric hash functions, and minimizes the feature gap with the help of a novel adaptive bipartite graph constructed on cross-domain data. By jointly optimizing asymmetric hash functions and the bipartite graph, not only can knowledge transfer be achieved but information loss caused by feature alignment can also be avoided. Meanwhile, to alleviate negative transfer, the intrinsic geometrical structure of single-domain data is preserved by involving a domain affinity graph. Extensive experiments on both single-domain and cross-domain benchmarks under different GITR subtasks indicate the superiority of our ATH method in comparison with the state-of-the-art hashing methods.  
\end{abstract}


\section{Introduction}
\IEEEPARstart{D}{ue} to the fast retrieval speed and low memory footprint, learning to hash \cite{hashingsurvey, GCNH, BDMFH, UBLH, ECF} has attracted extensive attention in different visual tasks over the past decades. The hashing approach aims to encode the data into compact binary codes so as to conduct fast retrieval based on hardware-level XOR operation. Recently, a large number of hashing methods have been proposed from different perspectives \cite{HUE,FCMH,LAKS,NAMVH,Labelself}. However, the known hashing approaches mainly rely on two assumptions: 1) sufficient training samples are required to learn discriminative hash functions; 2) query and retrieval samples come from the same domain with the same data distribution. In real-world scenarios, these assumptions are often easily violated. 
Specifically, the data available in the domain of interest may be inadequate for training, leading to data sparsity problem \cite{TH}. Moreover, there may exist an inevitable data distribution discrepancy between the query and retrieval domains. 
To tackle these drawbacks, a feasible solution is to integrate hashing approaches with Transfer Learning (TL) \cite{TCA}. TL aims to enhance the performance of target solver/learner on target domain by transferring the knowledge contained in different but related source domain \cite{ACE, MEDA}. As a result, the dependence on sufficient target domain data can be mitigated for target solver construction.

Inspired by TL, a few number of Transfer Hashing (TH) methods have been proposed recently \cite{TH,TAH,GTH,PWCF,DHLing}. The first TH method termed Transfer Hashing with Privileged Information (THPI) \cite{TH} marries hashing and transfer learning approaches to solve the data sparsity issue. Unlike THPI, an Optimal Projection Guided Transfer Hashing (GTH) \cite{GTH} seeks for a maximum likelihood estimation solution of hash functions of target and source domains. The newly-developed Probability Weighted Compact Feature learning (PWCF) \cite{PWCF} provides inter-domain correlation guidance to promote cross-domain retrieval accuracy, and Discriminative Hashing Learning (DHLing) \cite{DHLing} constructs a domain-invariant memory bank to achieve cross-domain alignment. The above-mentioned methods, however, are all homogeneous TH methods and cannot be directly applied for heterogeneous domains. Specifically, these methods assume that target and source domains share the same feature space with the same feature dimensionality. This assumption may not always hold since the features of the two domains may be heterogeneous or misaligned. For example, the query and retrieval samples may be of different resolutions and sizes, represented by hand-crafted features with different types and lengths (e.g., color-histogram \cite{CH}, SIFT \cite{SIFT}), or extracted by distinctive deep neural networks (e.g., VGG \cite{VGG}, ResNet \cite{ResNet}). In these cases, the known TH methods need an additional feature preprocessing (e.g, image scaling and dimension reduction) for feature alignment, which may cause information loss to a certain extent.

\begin{figure*}[!tb]
	\centering
	\includegraphics[scale=0.60,trim=60 110  45 170,clip]{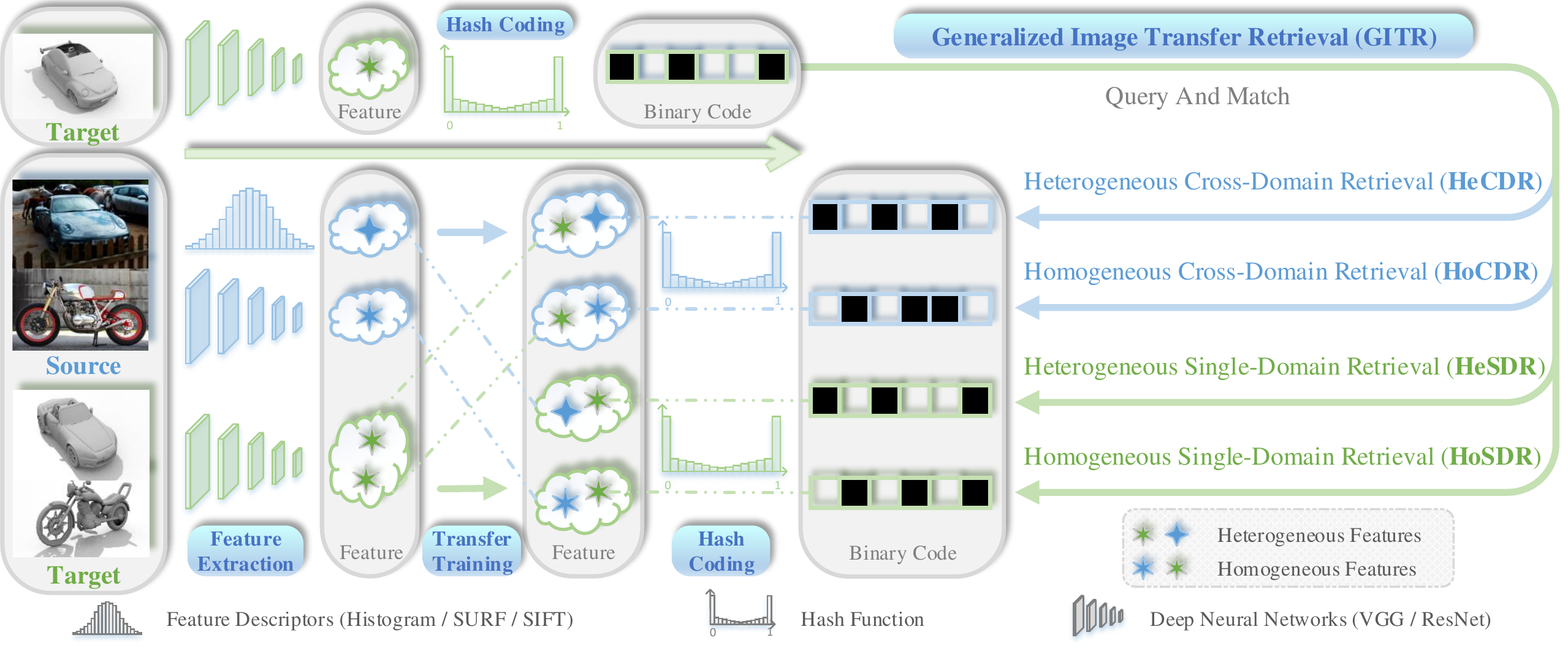}
	\caption{Illustration of the proposed GITR problem. Without loss of generality, we assume that the query samples come from the target domain. Given a query image, heterogeneous cross-domain retrieval (HeCDR) and homogeneous cross-domain retrieval (HoCDR) aim to find the most similar samples from the source domain, while heterogeneous single-domain retrieval (HeSDR) and homogeneous single-domain retrieval (HoSDR) tend to search the ones from the target domain. Here, ``He" means the features of query and retrieval sets are heterogeneous (i.e., using different feature descriptors/extractors), while ``Ho" indicates homogeneous features. Best viewed in colors.}
	\label{GITR}
\end{figure*}

In this paper, we first summarize a more challenging and practical transfer retrieval task termed \textit{Generalized Image Transfer Retrieval (GITR)}, in which the query and retrieval images may come from different domains and their features may be homogeneous or heterogeneous. Figure \ref{GITR} gives an illustration of the proposed GITR, which contains four different subtasks, including HeCDR, HoCDR, HeSDR, and HoSDR. Without any limitation on the feature spaces and domains of query and retrieval sets, this paper aims to develop a generalized framework to solve the complicated GITR problem. Specifically, given a query image observed in $\mathcal{F}_1$ feature space from $\mathcal{D}_1$ domain, we aim to retrieve the most similar images represented in $\mathcal{F}_2$ feature space from $\mathcal{D}_2$ domain, no matter whether these two domains and two feature spaces are the same or not. To achieve this goal, we propose an \textit{Asymmetric Transfer Hashing (ATH)} framework with its unsupervised/semi-supervised/supervised realizations. The proposed ATH contains three main components: a) asymmetric hashing learning, which characterizes domain distribution gap by the discrepancy of two asymmetric hash functions; b) cross-domain information interaction, which optimizes an adaptive bipartite graph for knowledge transfer and information fusion of cross-domain data; and c) domain structure preserving, which alleviates negative transfer by maintaining intrinsic geometrical structure of single-domain data based on domain affinity graph. 
The main contributions of this paper can be summarized as follows:
\begin{itemize}
	\item We extend the current learning-to-hash problem to a more general and challenging case, dubbed GITR, which makes no assumption or restrictions on the domains and feature spaces of query and retrieval sets. 
	\item To solve the GITR problem, we propose an ATH framework with its unsupervised/semi-supervised/supervised realizations. Specifically, ATH finds two asymmetric hash functions to encode the samples into discriminative hash codes, and learns an adaptive bipartite graph to characterize the similarity between cross-domain samples.
	\item To our best knowledge, ATH is the first heterogeneous transfer hashing framework that jointly optimizes the asymmetric hash functions and an adaptive bipartite graph in a unified learning objective, solving the two key problems in GITR simultaneously (see Fig. 2).
	\item Extensive experiments conducted on both single-domain and cross-domain benchmarks indicate that the known hashing methods tailored to a special feature space and domain cannot achieve satisfactory performance, while our ATH framework can work well in different scenarios.
\end{itemize}

The rest of this paper is organized as follows. The proposed GITR problem is illustrated in Section II where we also review some related works. Section III presents the proposed ATH framework with its unsupervised/semi-supervised/supervised realizations. In Section IV, we design an iterative
algorithm to optimize the objective of ATH and discuss the relationships between ATH and some related works. Section V
reports on the experimental results for evaluating the performance of
ATH. Conclusions and future work are covered in Section VI.

\section{Preliminaries}
\subsection{Generalized Image Transfer Retrieval}
Suppose that we have $n=n_s+n_t$ data samples from two domains $\mathcal{D}=\{\mathcal{D}_s, \mathcal{D}_t\}$, where source domain $\mathcal{D}_s$ and target domain $\mathcal{D}_t$ contain $n_s$ and $n_t$ samples, respectively. For each domain $\mathcal{D}_k$, $k\in\{s,t\}$, we have $\mathcal{D}_k=\{\mathbf{X}_k, \mathbf{Y}_k\}=\{(\mathbf{x}_{k,i}, \mathbf{y}_{k,i})\}_{i=1}^{n_k}$, where $\mathbf{x}_{k,i} \in \Re^{d_k}$ denotes the $i$-th sample of $\mathbf{X}_k \in \Re^{d_k\times n_k}$ with dimensionality $d_k$, and $\mathbf{y}_{k,i} \in \Re^{c}$ is the corresponding one-hot encoding label vector of $\mathbf{x}_{k,i}$. For simplicity, we assume that there are $c$ classes in both the source and target domains. The hash codes of data $\mathbf{X}_k$ are represented by $\mathbf{B}_k \in \Re^{r\times n_k}$, where $r$ is the length of binary codes. Moreover, $\mathbf{I}$, $\mathbf{1}$, and $\mathbf{0}$ denote the identity matrix, column vectors of all ones, and column vectors of all zeros with compatible size, respectively.

In the GITR problem, we do not require that the query and retrieval samples come from the same domain\footnote{Note that, the definition of transfer hashing problem is different from that of cross-modal hashing. Refer to \cite{TH, GTH, PWCF, DHLing} for more details.}. In addition, we do not limit that the feature spaces of the two domains are homogeneous. As a result, the assumption $d_s = d_t$ existed in previous TH methods may be violated. As illustrated in Figure \ref{GITR}, there exists domain consistency for HeSDR and HoSDR subtasks, and domain diversity for HeCDR and HoCDR subtasks. In addition, there exists feature homogeneity for HoSDR and HoCDR subtasks, and feature heterogeneity for HeSDR and HeCDR subtasks.

\subsection{Homogeneous Transfer Hashing}
\label{HTH}

The first TH method THPI \cite{TH} incorporates privileged information from the source domain into the target domain to assist hashing learning. 
Its objective function can be formulated as follows:
\begin{equation}
\begin{split}
&
\min_{\mathbf{B}_t, \mathbf{A}_t, \mathbf{A}_{sc}}
||\mathbf{E}||^2+\lambda_1||\mathbf{E}-\mathbf{A}_{sc}^T\mathbf{X}_{sc}||^2, \\
&\quad\ \ 
s.t.\quad \mathbf{A}_t^T\mathbf{A}_t=\mathbf{I},\ \mathbf{A}_{sc}^T\mathbf{A}_{sc}=\mathbf{I},
\end{split} 
\end{equation}
where $\mathbf{E}=\mathbf{B}_t-\mathbf{A}_t^T\mathbf{X}_t$, $\mathbf{X}_{sc} \in \Re^{d_s \times n_t}$ is $n_t$ selected samples from source domain $\mathcal{D}_s$, $\mathbf{A}_{t}\in \Re^{d_t\times r}$ and $\mathbf{A}_{sc}\in \Re^{d_s\times r}$ are orthogonal projection matrices for data $\mathbf{X}_t$ and $\mathbf{X}_{sc}$, respectively. As we can see, THPI utilizes data information of source domain to approximate the quantization errors of target domain. Due to $n_t \neq n_s$, THPI only uses $n_t$ source domain samples to conduct information interaction with $n_t$ target domain samples, which may weaken the transfer ability of target learner. In addition, how to select $n_t$ samples from the whole source domain remains to be determined.

Unlike THPI, GTH \cite{GTH} does not require to select a part of data from source domain for cross-domain information interaction. Its objective can be formulated as follows:
\begin{equation}
\begin{split}
&
\min_{\mathbf{B}_t, \mathbf{B}_s, \mathbf{A}_t, \mathbf{A}_s}
||\mathbf{M}^{\frac{1}{2}}\odot(\mathbf{A}_t-\mathbf{A}_s)||^2+\lambda_1||\mathbf{B}_t-\mathbf{A}_t^T\mathbf{X}_t||^2\\
&\quad +\lambda_2||\mathbf{B}_s-\mathbf{A}_s^T\mathbf{X}_s||^2, \ \ 
s.t.\ \  \mathbf{A}_t^T\mathbf{A}_t=\mathbf{I},\ \mathbf{A}_s^T\mathbf{A}_s=\mathbf{I},
\end{split} 
\end{equation}
where $\odot$ denotes element-wise product, $\mathbf{M}$ is an error matrix derived from maximum likelihood estimation, $\mathbf{A}_{s} \in \Re^{d_s\times r}$ and $\mathbf{A}_{t}\in \Re^{d_t\times r}$ are hashing projections for domain $\mathcal{D}_{s}$ and $\mathcal{D}_{t}$, respectively. 
GTH assumes that the similarity of images between target and source domains can be reversely characterized by discrepancy between hashing projections of the two domains. In other words, the weighted discrepancy between hashing projections $\mathbf{M}^{\frac{1}{2}}\odot(\mathbf{A}_t-\mathbf{A}_s)$ should be minimized to ensure that similar images from the target and source domains can be converted into similar hash codes. In this way, the whole knowledge in source domain can be used for target learner. 

Instead of minimizing the discrepancy between hashing projections, PWCF \cite{PWCF} designs a focal-triplet loss to promote cross-domain correlations and obtains good transfer performance.  Moreover, DHLing \cite{DHLing} further considers the problem of insufficient annotated source images and provides a semi-supervised approach. 
Nevertheless, all these TH methods \cite{TH,GTH,PWCF,DHLing} assume that the target and source domains share a homogeneous feature space with aligned features (i.e., $d_s=d_t=d$). This assumption may not always hold since the feature dimensionalities of retrieval and query samples may be misaligned. For example, the query and retrieval samples may be of distinctive resolutions and sizes, represented by hand-crafted features with different lengths, or extracted by different deep neural networks. In these cases, these methods cannot be directly applied to HeSDR and HeCDR subtasks (where $d_t \neq d_s$), which motivates us to develop a novel heterogeneous TH method to alleviate this problem.

\subsection{Asymmetric Hashing}
The earlier explorations on asymmetric hashing \cite{ASYH1, ASYH2} employ asymmetric distance metric for similarity preserving, where the hash codes of training and test sets are generated from the same hash function. 
In \cite{AsyHashing}, the authors have demonstrated that using different hash functions for query and retrieval sets can achieve better approximation of target similarity with shorter code lengths. Following this work, recent asymmetric hashing methods \cite{MAH,DSAH,DAPH,ADSH} tend to focus on functional space, where the distribution gap between training and test sets are characterized by the discrepancy of two distinct hash functions. In general, these methods adopt a so-called similarity-similarity difference function \cite{hashingsurvey} for similarity preservation:
\begin{equation}
\sum_{i}\sum_{j}\left|\left|f_1(\mathbf{x}_{i})f_2(\mathbf{x}_{j}) - r\mathbf{S}_{ij}\right|\right|_{}^2
\end{equation} 
where $\mathbf{S}$ is a pre-defined label matrix, $f_1(\cdot)$ and $f_2(\cdot)$ are two distinct hash functions. Technically, they approximate the similarity between samples $\mathbf{x}_i$ and $\mathbf{x}_j$ as the Hamming distance between $f_1(\mathbf{x}_i)$ and $f_2(\mathbf{x}_j)$. The more similar the two samples $\mathbf{x}_i$ and $\mathbf{x}_j$ are, the smaller the Hamming distance between $f_1(\mathbf{x}_i)$ and $f_2(\mathbf{x}_j)$ should be. Being different from these approaches which focus on single-domain retrieval, this paper utilizes the asymmetric hashing mechanism to solve the GITR problem, especially for its challenging HeSDR and HeCDR subtasks. More importantly, since there is no label information provided for the construction of $\mathbf{S}$ in the unsupervised scenario, we design a novel distance-similarity product function equipped with an adaptive bipartite graph to learn the potential semantic information from data.

\section{Methodology}

\subsection{Graph and Bipartite Graph}
Suppose that $\mathcal{G}=(\mathcal{V},\mathcal{E},\mathbf{W})$ is an undirected weighted graph with vertex set $\mathcal{V}(\mathcal{G})$, edge set $\mathcal{E}(\mathcal{G})$, and adjacent matrix $\mathbf{W}$, where each element of the real symmetric matrix $\mathbf{W}$ measures the affinity for a pair of vertices. In general, there are three main approaches for the construction of affinity graph: 1) \textit{nearest neighbor graph} \cite{Iscen17,JSH,Liu19}, which focuses on local neighborhood relationship of the data; 2) \textit{semantic graph} \cite{GER,Zhang17,LGCNH}, which utilizes prior semantic information of labels; 3) \textit{adaptive graph} \cite{RLRR,LCL19,LRAGE}, which learns adaptive graph structure from the data.
On the other hand, bipartite graph has been extensively explored in different applications \cite{LuoMM,IMC}, which can be defined as follows:
\begin{myDef}
	A graph $\mathcal{G}=(\mathcal{V},\mathcal{E},\mathbf{W})$ is called bipartite if $\mathcal{V}=\mathcal{V}_s\cup\mathcal{V}_t$ and $\mathcal{V}_s\cap\mathcal{V}_t=\emptyset$, such that every edge in $\mathcal{E}$ connects one vertex from $\mathcal{V}_s$ to one vertex from $\mathcal{V}_t$. \label{definition1}
\end{myDef}
Note that, the affinity matrix $\mathbf{W}\in \Re^{\mid\mathcal{V}_s\mid\times \mid\mathcal{V}_t\mid}$ of a bipartite graph $\mathcal{G}=(\mathcal{V},\mathcal{E},\mathbf{W})$ will be asymmetric if $\mid\mathcal{V}_s\mid\neq\mid\mathcal{V}_t\mid$.

\begin{figure}[!tb]
	\centering
	\includegraphics[scale=0.37,trim=250 150  120 70,clip]{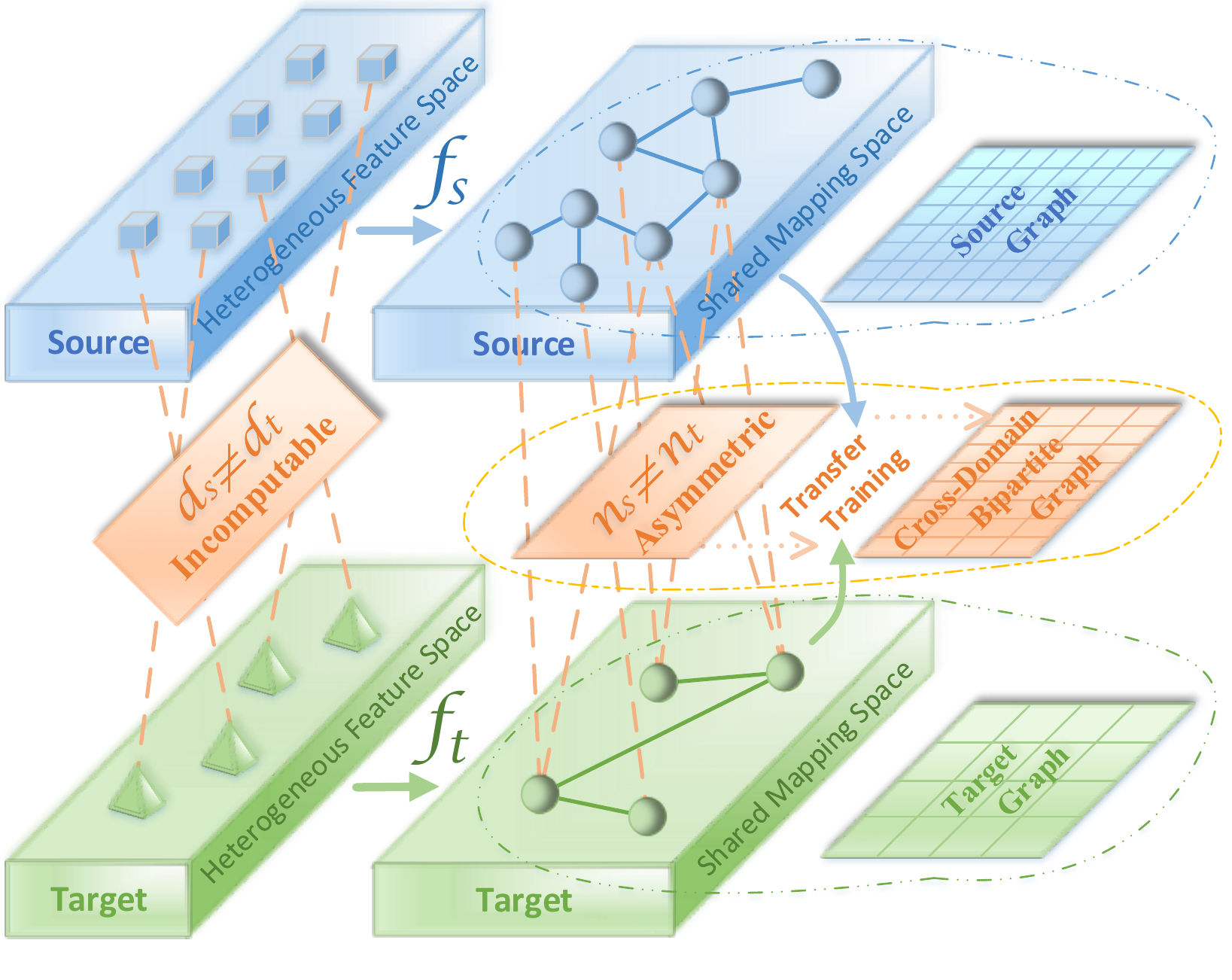}
	\caption{Technically, the GITR faces two dilemmas: $d_s \neq d_t$ and $n_s \neq n_t$. For $d_s \neq d_t$, ATH learns two distinct hash functions $f_s$ and $f_t$ to transform the heterogeneous data into an alignment mapping space. For $n_s \neq n_t$, ATH optimizes an adaptive bipartite graph to integrate the whole cross-domain data for transfer training. Moreover, in order to alleviate negative transfer, ATH preserves the intrinsic geometrical structure of single-domain data by involving a domain affinity graph. Best viewed in colors.} 
	\label{core_idea}
\end{figure}

\subsection{Unsupervised ATH}
Figure \ref{core_idea} illustrates the core idea of the proposed ATH framework, which contains three main parts including asymmetric hashing learning, cross-domain information interaction, and domain structure preserving.  In this section, we first consider an unsupervised ATH (ATH\_U), where neither the source domain nor the target domain contain labeled samples as the application scenarios in \cite{TH, GTH}.

\subsubsection{Asymmetric Hashing Learning}
In ATH, we exploit two distinct hash functions ${f}_t$ and ${f}_s$ to encode target and source samples separately, as follows:
\begin{equation}
\begin{split}
&
\min_{\mathbf{B}_k, f_k} \mathcal{T}_1 = 
\sum_{k \in \{s, t\}}\alpha_k||\mathbf{B}_k-{f}_k(\mathbf{X}_k)||_F^2,  \\
&\ 
s.t.\quad \mathbf{B}_k\in \{-1,1\}^{r\times n_k},  
\mathbf{B}_k\mathbf{1}=\mathbf{0},
\end{split} \label{AHL}
\end{equation}
where $\alpha_k \in \{\alpha_s, \alpha_t\}$ is a balance parameter, and ${f}_k \in \{{f}_s, {f}_t\}$ is the hash function of domain $\mathcal{D}_k$. We impose a constraint $\mathbf{B}_k\mathbf{1}=\mathbf{0}$ to make each bit to have an equal chance to be $1$ or $-1$, thereby maintaining the most information of data. The designed optimization objective (\ref{AHL}) aims to model the fitting error between the binary code $\mathbf{B}_k$ and hash mapping ${f}_k(\mathbf{X}_k)$ for each domain $\mathcal{D}_k$. Note that, $f_k(\cdot)$ can be any suitable mapping function, such as linear function and kernel function. For ATH\_U, we just exploit a simple yet effective linear function:
\begin{equation}
f_k(\mathbf{x}_{k,i}) = \mathbf{A}_{k}^T\mathbf{x}_{k,i},
\end{equation}
where $\mathbf{A}_k \in \{\mathbf{A}_s, \mathbf{A}_t\}$ is a hashing projection. Most notably, $\mathbf{A}_{s} \in \Re^{d_s\times r}$ and $\mathbf{A}_{t}\in \Re^{d_t\times r}$ can be of the same magnitude or not. If $d_s = d_t$, it can be used for HoSDR and HoCDR. Otherwise, it is tailored to HeSDR and HeCDR.

\subsubsection{Cross-Domain Information Interaction}
We consider conducting direct knowledge transfer and fusion between target and source domains, regardless of whether these two domains share the same feature space or not. Obviously, it is infeasible to minimize the errors between $\mathbf{A}_s$ and $\mathbf{A}_t$ as in \cite{GTH} since the cross-domain features may be misaligned.
According to Definition \ref{definition1}, we observe that if we regard the source and target data as a whole dataset, on which we can naturally construct a bipartite graph $\mathcal{G}_{st}=(\mathcal{V}_{st},\mathcal{E}_{st},\mathbf{W}_{st})$ where $\mathcal{V}_{st}=\mathbf{X}_s\cup\mathbf{X}_t$ and $\mathbf{W}_{st}\in \Re^{n_s\times n_t}$. Nevertheless, how to determine the edge set $\mathcal{E}_{st}$ and edge weight $\mathbf{W}_{st}$ becomes a knotty problem. Note that, we cannot apply a nearest neighbor bipartite graph in HeSDR and HeCDR subtasks since the distances between heterogeneous features of cross-domain data cannot be calculated directly, as shown in Figure \ref{core_idea}. To mitigate this problem, we propose to learn an adaptive bipartite graph from data to characterize the affinities of cross-domain samples, by the following distance-similarity product function:
\begin{equation}
\begin{split}
&\min_{\mathbf{W}_{st}, f_k} \mathcal{T}_2 = 
\lambda\sum_{i=1}^{n_s}\sum_{j=1}^{n_t}\mathbf{F}_{st,ij}\mathbf{W}_{st,ij} + \gamma||\mathbf{W}_{st}||_F^2,  \\
&\quad\quad\  \mathit{s.t.} \quad \forall_i, \quad  \mathbf{W}_{st,i:}\geq 0, \quad \mathbf{W}_{st,i:}^T\mathbf 1=1
\end{split}
\label{UDA}
\end{equation}
where $\lambda$ and $\gamma$ are two hyper-parameters, $\mathbf{W}_{st,i:}$ is the $i$-th row of $\mathbf{W}_{st}$, and $\mathbf{F}_{st,ij}=||f_s(\mathbf{x}_{s,i})-f_t(\mathbf{x}_{t,j})||^2$ models the distances (or fitting errors) between hash mappings of cross-domain samples. The value of $\mathbf{W}_{st,ij}$ can be viewed as the probability that two cross-domain samples $\mathbf{x}_{s,i}$ and $\mathbf{x}_{t,j}$ belong to the same class. Therefore, we make the constraint $\mathbf{W}_{st,i:}\geq 0$. In addition, the constraint $\mathbf{W}_{st,i:}\mathbf 1=1$ is imposed to avoid a trivial case where all elements of $\mathbf{W}_{st,i:}$ are zero, and the regularization term $\gamma||\mathbf{W}_{st}||_F^2$ provides the probability prior of uniform distribution. 
As we can see, the designed distance-similarity product function (\ref{UDA}) enables the solver to simultaneously learn hash functions and affinity relationship of cross-domain samples, regardless of whether the features of the two domains are misaligned or not. This optimization objective stems from a straightforward intuition that the more similar two cross-domain samples (i.e., the larger $\mathbf{W}_{st,ij}$) are, the less difference the corresponding hash mappings (i.e., the smaller $\mathbf{F}_{st,ij}$) should be. The joint optimization of asymmetric hash functions and bipartite graph promotes knowledge fusion between cross-domain data and makes it possible to conduct transfer training in heterogeneous domains.

\subsubsection{Domain Structure Preserving} When performing transfer training, negative transfer becomes a problem worthy of careful consideration. To mitigate this problem, we preserve the intrinsic geometrical structure of single-domain data by involving a domain affinity graph. Specifically, for each domain $\mathcal{D}_k$, we first construct a nearest neighbor graph $\mathcal{G}_{k}=(\mathcal{V}_{k},\mathcal{E}_{k},\mathbf{W}_{k})$, where $\mathcal{V}_{k}=\mathbf{X}_k$ and $\mathcal{E}_{k}$ is defined as:
\begin{equation}
\mathcal{E}_{k}=\left\{<\mathbf{x}_{k,i}, \mathbf{x}_{k,j}>\ |\ i\in \mathcal{N}_{\eta_k}(j) \cup j\in \mathcal{N}_{\eta_k}(i)\right\}
\end{equation}
where $\mathcal{N}_{\eta_k}(i)$ denotes the set of $\eta_k$ nearest neighbors of $\mathbf{x}_{k,i}$. With the defined edge set $\mathcal{E}_{k}$, we can simply set the edge weight $\mathbf{W}_{k}\in \Re^{n_k\times n_k}$ as follows:
\begin{equation}
\mathbf{W}_{k,ij}=\left\{
\begin{array}{ll}
1,&\mathrm{if}\ <\mathbf{x}_{k,i}, \mathbf{x}_{k,j}> \ \in\  \mathcal{E}_{k}, \\
0,&\mathrm{otherwise}
\end{array}
\right. 
\end{equation}
Then, we design the following optimization objective to force the similar samples of each domain to be as close as possible in the corresponding mapping space:
\begin{equation}
\min_{f_k} \mathcal{T}_3 = 
\sum_{k \in \{s, t\}}\frac{\beta_k}{2}\sum_{i=1}^{n_k}\sum_{j=1}^{n_k}\mathbf{F}_{k,ij}\mathbf{W}_{k,ij},  \label{DSP}
\end{equation}
where $\beta_k \in \{\beta_s, \beta_t\}$ is a balance parameter, and $\mathbf{F}_{k,ij}=||f_k(\mathbf{x}_{k,i})-f_k(\mathbf{x}_{k,j})||^2$ models the fitting errors between hash mappings of single-domain samples.

\subsubsection{Overall Objective Function}
The overall objective function of the proposed ATH\_U is formulated as follows:
\begin{equation}
\begin{split}
&\quad\ \ 
\min_{\mathbf{B}_k, \mathbf{W}_{st}, f_k} \mathcal{J} = \mathcal{T}_1 + \mathcal{T}_2 + \mathcal{T}_3  \\
&
\mathit{s.t.}\quad \mathbf{B}_k\in \{-1,1\}^{r\times n_k},  
\quad \mathbf{B}_k\mathbf{1}=\mathbf{0}, \\ 
&\quad\ \  \forall_i, \quad  \mathbf{W}_{st,i:}\geq 0, \quad \mathbf{W}_{st,i:}^T\mathbf 1=1
\end{split} \label{ATH_obj}
\end{equation}
The joint objective function (\ref{ATH_obj}) can be optimized by a well-designed iterative algorithm, as shown in Section \ref{optimization}.

\begin{algorithm}[!tb]
	\label{algorithm}
	\caption{Optimization algorithm for ATH\_U}
	\LinesNumbered
	\KwIn{Training sets $\mathbf{X}_k \in \{\mathbf{X}_s, \mathbf{X}_t\}$, maximum iterations $Ite$, parameters $\alpha_k$, $\beta_k$, $\lambda$, $r$.}
	\KwOut{Hashing projections $\mathbf{A}_k \in \{\mathbf{A}_s, \mathbf{A}_t\}$.}
	
	Construct affinity matrices $\mathbf{W}_k \in \{\mathbf{W}_s, \mathbf{W}_t\}$; 
	
	Initialize $\mathbf{A}_k$ by PCA projections;
	
	Initialize $\mathbf{B}_k$ by $\mathbf{B}_k = sign(\mathbf{A}_k^T\mathbf{X}_k)$;
	
	\While{converge or reach maximum iterations $Ite$}{
		Update $\mathbf{A}_k$ according to (\ref{A_sol});
		
		Update $\mathbf{M}_k$ by $\mathbf{M}_k = \alpha_k\mathbf{A}_k^T\mathbf{X}_k$;
		
		\For{$i = 1:r$}{			
			Update $\mathbf{B}_{k,ij}$ according to (\ref{B_sol});
		}	
	
		\For{$i = 1:n_s$}{	
			Update $\mathbf{W}_{st,i}$ according to (\ref{WST_solution});
		}
	}
\end{algorithm}

\subsection{Semi-supervised ATH} 
This section provides a semi-supervised realization of ATH termed ATH\_M, where all samples in the source domain are well annotated as the application scenarios in \cite{PWCF, DHLing}. The difference of objective functions between ATH\_U and ATH\_M lies in the construction of source graph. Unlike ATH\_U, ATH\_M adopts a semantic graph for the source domain. Specifically, ATH\_M defines the edge set $\mathcal{E}_{s}$ of source graph $\mathcal{G}_{s}=(\mathcal{X}_{s},\mathcal{E}_{s},\mathbf{W}_{s})$ as follows:
\begin{equation}
\mathcal{E}_{s}=\left\{<\mathbf{x}_{s,i}, \mathbf{x}_{s,j}>\ |\  \mathbf{y}_{s,i}=\mathbf{y}_{s,j}\right\} \label{edge_M}
\end{equation}
The edge weight setting of ATH\_M is similar to that of ATH\_U. As can be seen from Eq. \ref{edge_M}, ATH\_M exploits semantic information of the source domain to perform transfer training, which promotes the performance of hash encoders.

\subsection{Supervised ATH and Kernel Extension}
This section presents a supervised realization of ATH dubbed ATH\_S, which is tailored to the case where all samples in the source and target domains are well labeled. With the available label information, we do not need to optimize an adaptive bipartite graph for cross-domain knowledge transfer as in Eq. \ref{UDA}. Instead, ATH\_S proactively constructs a semantic bipartite graph and sets its edge set $\mathcal{E}_{st}$ as:
\begin{equation}
\mathcal{E}_{st}=\left\{<\mathbf{x}_{s,i}, \mathbf{x}_{t,j}>\ |\  \mathbf{y}_{s,i}=\mathbf{y}_{t,j}\right\} \label{edge_st}
\end{equation}
In this way, semantic similarity of cross-domain data can be used to conduct domain alignment and improve the transferability of our model. As for the source and target graphs $\mathcal{G}_{k}=(\mathbf{X}_k,\mathcal{E}_{k},\mathbf{W}_{k})$, we also utilize semantic information to construct the edge set $\mathcal{E}_{k}$ for each domain $\mathcal{D}_k$ as follows:
\begin{equation}
\mathcal{E}_{k}=\left\{<\mathbf{x}_{k,i}, \mathbf{x}_{k,j}>\ |\  \mathbf{y}_{k,i}=\mathbf{y}_{k,j}\right\} \label{edge_k}
\end{equation}

As mentioned in Eq. \ref{AHL}, we can use a kernel function for realizing $f_k$. To explore the performance of kernel function, we further present a kernel extension of ATH\_S named ATH\_K, of which the hash function can be formulated as: 
\begin{equation}
\hat{f}_k(\mathbf{x}_{k,i}) = \mathbf{A}_{k}^T\phi(\mathbf{x}_{k,i}),
\end{equation}
where $\phi(\mathbf{x}_{k,i})$ is a $m_k \in \{m_s, m_t\}$-dimensional column vector obtained by Gaussian kernel 
\begin{small}
\begin{equation}
\phi(\mathbf{x}_{k,i}) = \left[\exp \left(\frac{||\mathbf{x}_{k,i} - \mathbf{\chi}_{k,1}||^2}{\sigma}\right), \cdots, \exp \left(\frac{||\mathbf{x}_{k,i} - \mathbf{\chi}_{k,m_k}||^2}{\sigma}\right)\right]^T
\end{equation}
\end{small}where $\{\mathbf{\chi}_{k,j}\}_{j=1}^{m_k}$ are $m_k$ anchor points randomly selected from domain $\mathcal{D}_k$, and $\sigma$ is the kernel width \cite{SDH}.

\section{Discussion}

\subsection{Optimization}
\label{optimization}
Due to space limitation, we only provide the optimization algorithm for ATH\_U. The optimization approaches of ATH\_M, ATH\_S, and ATH\_K are similar to that of ATH\_U. Specifically, our iterative algorithm updates one variable at a time by fixing the others, as listed in Algorithm \ref{algorithm}. 

\textbf{$\mathbf{A}_k$-Step.} The subproblem of $\mathbf{A}_k\in\{\mathbf{A}_s,\mathbf{A}_t\}$ is equal to the following unconstrained minimization problem: 
\begin{equation}
	\begin{split}
	&\min_{\mathbf{A}_k}\ 
	\alpha_k||\mathbf{B}_k-\mathbf{A}_k^T\mathbf{X}_k||_F^2+\beta_ktr(\mathbf{A}_k^T\mathbf{X}_k \mathbf{L}_k\mathbf{X}_k^T\mathbf{A}_k)\\
	&\ \ +\lambda tr(-2\mathbf{A}_k^T\mathbf{X}_k\mathbf{\widetilde{W}}_{st}\mathbf{\widetilde{X}}_k^T\mathbf{\widetilde{A}}_k+\mathbf{A}_k^T\mathbf{X}_k \mathbf{\widetilde{D}}_{st} \mathbf{X}_k^T\mathbf{A}_k)\\
	&=tr(\mathbf{A}_k^T(\mathbf{X}_k(\alpha_k\mathbf{I}+\beta_k \mathbf{L}_k + \lambda \mathbf{\widetilde{D}}_{st})\mathbf{X}_k^T)\mathbf{A}_k\\
	&\ \ -2\mathbf{A}_k^T\mathbf{X}_k(\alpha_k\mathbf{B}_k^T+\lambda\mathbf{\widetilde{W}}_{st}\mathbf{\widetilde{X}}_k^T\mathbf{\widetilde{A}}_k)),
	\end{split} \label{A_sub}
\end{equation}
where $\mathbf{\widetilde{D}}_{st}\in\Re^{n_k\times n_k}$ is a diagonal matrix, $\mathbf{L}_{k} \in \{\mathbf{L}_s, \mathbf{L}_t\}$ is the graph Laplacian matrix, $\mathbf{L}_{k} = \mathbf{D}_{k}- \mathbf{W}_{k}$, and $\mathbf{D}_{k}\in\Re^{n_k\times n_k}$ is a diagonal matrix with $\mathbf{D}_{k, ii} = \sum_j \mathbf{W}_{k, ij}$. 
In (\ref{A_sub}), if $k=s$, $\mathbf{\widetilde{D}}_{st, ii}=\sum_j \mathbf{W}_{st, ij}$, $\mathbf{\widetilde{W}}_{st}=\mathbf{W}_{st}$, $\mathbf{\widetilde{X}}_k=\mathbf{X}_t$, and $\mathbf{\widetilde{A}}_k=\mathbf{A}_t$. Otherwise, $\mathbf{\widetilde{D}}_{st, jj}=\sum_i \mathbf{W}_{st, ij}$, $\mathbf{\widetilde{W}}_{st}=\mathbf{W}_{st}^T$, $\mathbf{\widetilde{X}}_k=\mathbf{X}_s$, and $\mathbf{\widetilde{A}}_k=\mathbf{A}_s$. By setting the derivative of (\ref{A_sub}) with respect to $\mathbf{A}_k$ to zero, the optimal solution of $\mathbf{A}_k$ subproblem can be obtained by\footnote{In practice, we generally add a regularization as $\mathbf{\widetilde{\Phi}}_k = \mathbf{\Phi}_k+\varphi \mathbf{I}$ ($\varphi>0$) to solve the singularity problem if it exists.}:
\begin{equation}
\mathbf{A}_k=\mathbf{\Phi}_k^{-1}(\alpha_k\mathbf{X}_k\mathbf{B}_k^T+\lambda\mathbf{X}_k\mathbf{\widetilde{W}}_{st}\mathbf{\widetilde{X}}_k^T\mathbf{\widetilde{A}}_k),   \label{A_sol}
\end{equation}
where $\mathbf{\Phi}_k = \mathbf{X}_k(\alpha_k\mathbf{I}_{n_k}+\beta_k\mathbf{L}_k+\lambda\mathbf{\widetilde{D}}_{st})\mathbf{X}_k^T$.

\textbf{$\mathbf{B}_k$-Step.} The subproblem of $\mathbf{B}_k \in\{\mathbf{B}_s,\mathbf{B}_t\}$ is equal to the following non-convex problem with discrete constraints:
\begin{equation}
\max_{\mathbf{B}_k\in \{-1,1\}^{r\times n_k},  
	\mathbf{B}_k\mathbf{1}=\mathbf{0}} tr(\mathbf{B}_k^T\mathbf{M}_k),
\end{equation}
where $\mathbf{M}_k = \alpha_k\mathbf{A}_k^T\mathbf{X}_k$. The optimal solution of this maximization problem can be directly obtained by sorting the elements of each row of $\mathbf{M}_k$:
\begin{equation}
\mathbf{B}_{k, ij}=\left\{
\begin{array}{lr}
1, \qquad  \quad \mathrm{if}\ \  j\in \tau(i),& \\
-1,\qquad  \ \mathrm{otherwise}&
\end{array}
\right.  \label{B_sol}
\end{equation}
where $\tau(i)$ indicates an index set of the first $n_k/2$ maximal elements of $\mathbf{M}_{k,i:}$, and $\mathbf{M}_{k,i:}$ indicates the $i$-th row of $\mathbf{M}_k$.

\textbf{$\mathbf{W}_{st}$-Step.} To solve the subproblem of $\mathbf{W}_{st}$, we need to address $n_s$ decoupled independent problems as follows:
\begin{equation}
\begin{split}
&\ \  \min_{\mathbf{W}_{st,i:}}  ||\mathbf{W}_{st,i:}+\frac{\lambda}{2\gamma_i}\mathbf{F}_{st,i:}||^2,  \\
& \mathit{s.t.} \quad  \mathbf{W}_{st,i:}\geq 0, \quad \mathbf{W}_{st,i:}^T\mathbf 1=1
\end{split}
\label{WST1}
\end{equation}
For problem (\ref{WST1}), we have the following theorem:
\begin{myThe}
	Let the number of nonzero elements in $\mathbf{W}_{st,i:}$ be $\eta \in \{1,...,n_t\}$. Then the optimal solution of (\ref{WST1}) is:
	\begin{equation}
	\mathbf{W}_{st,i:} = \left(\frac{1+\sum_{j=1}^{\eta}\frac{\lambda}{2\gamma_i}\mathbf{\widetilde{F}}_{st,ij}}{\eta}\mathbf{1}-\frac{\lambda}{2\gamma_i}\mathbf{F}_{st,i:}\right)_{+}
	\label{WST_solution}
	\end{equation}
	where the operator $(\cdot)_+$ turns the negative elements to zero while keeps the rest, and the elements of $\mathbf{\widetilde{F}}_{st,i:}$ are those of $\mathbf{F}_{st,i:}$ but with the ascending order. 
\end{myThe}

\textbf{Proof.} The Lagrangian function of the $\mathbf{W}_{st}$ subproblem can be formulated as follows:
\begin{equation}
\begin{split}
\mathcal{L}= \frac{1}{2}||\mathbf{W}_{st,i:}+\frac{\lambda}{2\gamma_i}\mathbf{F}_{st,i:}||^2-\varphi(\mathbf{W}_{st,i:}^T\mathbf 1-1)-\mathbf{u}^T \mathbf{W}_{st,i:} 
\end{split}
\label{s1}
\end{equation}
where $\varphi$ and $\mathbf{u}$ are Lagrangian multipliers. By taking the derivate of $\mathcal{L}$ with respect to $\mathbf{W}_{st,i:}$ and setting it zero, with the KKT condition, we obtain:
\begin{equation}
\begin{split}
&\frac{\partial \mathcal{L}}{\partial \mathbf{W}_{st,i:}}= \mathbf{W}_{st,i:}+\frac{\lambda}{2\gamma_i}\mathbf{F}_{st,i:}-\varphi\mathbf 1-\mathbf{u}=0;\\ 
&\qquad \mathbf{W}_{st,i:}^T\mathbf 1-1=0; \quad \mathbf{u}^T \mathbf{W}_{st,i:}=0  
\end{split}
\label{s2}
\end{equation}
For the third equation, if $\mathbf{W}_{st,ij} > 0$, then we have $\mathbf{u}_j=0$. Therefore, the solution of $\mathbf{W}_{st,i:}$ is:
\begin{equation}
\begin{split}
\mathbf{W}_{st,i:}= \left(\varphi\mathbf 1- \frac{\lambda}{2\gamma_i}\mathbf{F}_{st,i:}\right)_+ 
\end{split}
\label{s3}
\end{equation}
Since $\mathbf{W}_{st,i:}$ has $\eta$ nonzero elements and $\mathbf{W}_{st,i:} \geq 0$, we have:
\begin{equation}
\begin{split}
\varphi - \frac{\lambda}{2\gamma_i}\mathbf{\widetilde{F}}_{st,i\eta} > 0 \quad \bigcap \quad \varphi - \frac{\lambda}{2\gamma_i}\mathbf{\widetilde{F}}_{st,i(\eta+1)} \leq 0 
\end{split}
\label{s4}
\end{equation}
According to $\mathbf{W}_{st,i:}^T\mathbf 1=1$ and Eq. \ref{s3}, we obtain:
\begin{equation}
\begin{split}
&\sum_{j=1}^{\eta} \left(\varphi-\frac{\lambda}{2\gamma_i}\mathbf{\widetilde{F}}_{st,ij}\right) = 1 \Rightarrow  \varphi = \frac{1+\sum_{j=1}^{\eta}\frac{\lambda}{2\gamma_i}\mathbf{\widetilde{F}}_{st,ij}}{\eta}
\end{split} 
\label{s5}
\end{equation}
Substitute Eq. \ref{s5} into Eq. \ref{s3}, and we can obtain the same solution as stated in Theorem 1. $\hfill\square$

\begin{table*}[!tb]
	\setlength{\abovecaptionskip}{-0cm}
	\setlength{\belowcaptionskip}{-0cm}
	\scriptsize
	\centering
	\setlength{\tabcolsep}{3.0mm}
	\caption{MAP (\%) results of different methods with varying code lengths from 16 to 128 on single-domain datasets.}
	\begin{tabular}{lcccccc|cccccc}
		\hline	
		\hline
		\multirow{2}*{Method} 		&\multicolumn{1}{|c}{16} &32 &48 &64 &96 &128 &16 &32 &48 &64 &96 &128\\		
		\cline{2-13}	
		& \multicolumn{6}{|c|}{SUN397 \texttt{\{train 106,953:test 1,800\}}} 
		& \multicolumn{6}{c}{CIFAR-10 \texttt{\{train 59,000:test 1,000\}}}  	\\
		\hline
		ITQ 
		&\multicolumn{1}{|c}{7.25} &9.57 &10.07 &10.69 &12.33  &12.70 & 15.22 &15.97 &16.10 &16.21  &16.72  &16.88\\
		SP 
		&\multicolumn{1}{|c}{8.68} &10.55 &11.60 &12.98 &14.14 &14.74  &16.23  &16.51 &16.98  &17.20 &17.61  &17.85 \\
		DPLM 
		&\multicolumn{1}{|c}{11.11} &13.71 &16.49 &\textcolor{blue}{17.81} &\textcolor{blue}{19.35} &\textcolor{blue}{19.88}  & 16.83    & \textcolor{blue}{18.31}    & \textcolor{blue}{19.32}    & \textcolor{blue}{19.50}    & \textcolor{blue}{20.13}    & \textcolor{blue}{20.17} \\
		SHSR 
		&\multicolumn{1}{|c}{\textcolor{blue}{12.40}} &\textcolor{blue}{16.31} &\textcolor{blue}{18.03} &16.76 &18.01 &17.87 &\textcolor{blue}{17.35} &17.05 &16.85 &16.47 &15.75 &15.36  \\
		GTH 
		&\multicolumn{1}{|c}{8.04} &11.31 &12.93 &13.51 &14.41 &15.36 &15.07    &16.23    &16.53    &16.77    &17.12    &16.40 \\
		ATH\_U  
		&\multicolumn{1}{|c}{\textcolor{red}{13.87}} &\textcolor{red}{17.47} &\textcolor{red}{19.14} &\textcolor{red}{19.89} &\textcolor{red}{22.09} &\textcolor{red}{22.90} &\textcolor{red}{18.01} &\textcolor{red}{19.54} &\textcolor{red}{20.69} &\textcolor{red}{20.82} &\textcolor{red}{20.91} &\textcolor{red}{21.02}  \\
		\hline
		\hline
	\end{tabular}
	\label{Table1}
\end{table*}

\subsection{Parameter Setting of $\gamma$}
In this section, we show how to determine the value of parameter $\gamma$. Combining Eq. \ref{s4} and \ref{s5}, we can obtain:
\begin{small}
	\begin{equation}
	\begin{split}
	&\frac{\lambda \eta}{2}\mathbf{\widetilde{F}}_{st,i\eta}-\frac{\lambda}{2}\sum_{j=1}^{\eta}\mathbf{\widetilde{F}}_{st,ij} < {\gamma_i}  \leq  \frac{\lambda \eta}{2}\mathbf{\widetilde{F}}_{st,i(\eta+1)} - \frac{\lambda}{2}\sum_{j=1}^{\eta}\mathbf{\widetilde{F}}_{st,ij}
	\end{split} 
	\label{s6}
	\end{equation}
\end{small}
In oder to make the optimal solution of $\mathbf{W}_{st,i:}$ contain $\eta$ nonzero elements, we can set $\gamma_i$ as:
\begin{equation}
\begin{split}
& {\gamma_i}  =  \frac{\lambda \eta}{2}\mathbf{\widetilde{F}}_{st,i(\eta+1)} - \frac{\lambda}{2}\sum_{j=1}^{\eta}\mathbf{\widetilde{F}}_{st,ij}
\end{split} 
\label{s7}
\end{equation}
As a result, we do not need to turn the value of parameter $\gamma$. We can set its value as the average of $\gamma_i$ for $i\in\{1,2,...,n_s\}$:
\begin{equation}
\begin{split}
& {\gamma}  = \frac{1}{n_s} \sum_{i=1}^{n_s}\left(\frac{\lambda \eta}{2}\mathbf{\widetilde{F}}_{st,i(\eta+1)} - \frac{\lambda}{2}\sum_{j=1}^{\eta}\mathbf{\widetilde{F}}_{st,ij}\right)
\end{split} 
\label{s8}
\end{equation}

\subsection{Connections to Previous Methods}
We analyze the connections and differences between the proposed ATH and some related hashing methods, including anchor-graph-based hashing \cite{Liu11, JSH, DAGH}, asymmetric hashing \cite{ADSH,NAMVH,MAH}, and transfer hashing \cite{TH, GTH, PWCF, DHLing}.

The anchor-graph-based hashing methods \cite{ Liu11, JSH, DAGH} aim to perform fast graph construction for large-scale image retrieval. In fact, anchor graph can be viewed as a special bipartite graph. Unlike these methods focusing on single-domain image retrieval, we utilize the bipartite graph to integrate cross-domain information for transfer hashing. Moreover, these methods are all two-step approaches, which first pre-construct an anchor graph through clustering and then use it to learn hash functions. Nevertheless, our proposed ATH can learn the bipartite graph and hash functions simultaneously in a unified framework.

The asymmetric hashing methods \cite{ADSH,NAMVH,MAH} emphasize achieving similarity approximation with short binary codes. To achieve this goal, they usually adopt the so-called similarity-similarity difference function \cite{hashingsurvey} for similarity preservation. Unlike these approaches, the proposed ATH aims to learn distinguish hash functions with unaligned inputs for the source and target domains by incorporating the asymmetric hashing mechanism. Therefore, ATH can solve the transfer hashing problem, especially for the HeCDR task with misaligned features. Moreover, ATH constructs a distance-similarity product function \cite{hashingsurvey} for similarity preserving, which helps to jointly optimize the bipartite graph and hash functions.

The transfer hashing methods \cite{GTH, PWCF, DHLing} are proposed to deal with cross-domain retrieval task. These methods find either a shared hash function or two different hash functions with aligned inputs for source and target domains. As a result, they cannot be directly applied in HeSDR and HeCDR tasks, since the feature dimensionalities of source and target domains in these two tasks are inconsistent. The proposed ATH can handle this problem by learning asymmetric hash functions. Moreover, the transfer hashing method \cite{TH} need to select a part of source samples to integrate information with the target samples, while our ATH can perform cross-domain information interaction for all source and target sample through the learning of a bipartite graph.

\subsection{Time Complexity}
As listed in Algorithm \ref{algorithm}, the main computation burden of ATH\_U comes from iteratively updating $\mathbf{A}_k$, $\mathbf{B}_k$, and $\mathbf{W}_{st}$. 
The main computation cost for optimizing $\mathbf{A}_k$ and $\mathbf{B}_k$ are $\mathcal{O}(d_k^3)$ and $\mathcal{O}(rn_klogn_k)$ respectively. And we need $\mathcal{O}(n_sn_tlogn_t)$ for the optimization of $\mathbf{W}_{st}$. In transfer hashing task, there usually exists $n_t\ll n_s$. Therefore, the computation cost of our algorithm is acceptable.

\begin{table*}[!tb]
	\scriptsize
	\setlength{\abovecaptionskip}{-0cm}
	\setlength{\belowcaptionskip}{-0cm}
	\centering
	\setlength{\tabcolsep}{1.8mm}
	\caption{MAP (\%) results of different methods with varying code lengths from 16 to 128 on the HoSDR subtask.}
	\begin{tabular}{l|cccccc|cccccc|cccccc}
		\hline
		\hline
		\multirow{2}*{Method} 
		& {16}   & {32}   & {48}   & {64}   & {96}   & {128}  
		& {16}   & {32}   & {48}   & {64}   & {96}   & {128} 
		& {16}   & {32}   & {48}   & {64}   & {96}   & {128}   \\
		\cline{2-19}
		& \multicolumn{6}{c|}{{P $\rightarrow$ A}}  & \multicolumn{6}{c|}{{R $\rightarrow$ C}} & \multicolumn{6}{c}{{P $\rightarrow$ C}}\\
		\hline	
		ITQ            
		& 14.40  & 25.21  & 28.25  & 30.88  & 33.38  & 34.21   
		& 8.36  & 12.41  & 14.81  & 16.24  & 17.74  & 18.46  
		& 8.18  & 12.34  & 14.34  & 15.39  & 17.05  & 17.54     \\
		SP          
		& \textcolor{black}{18.10}  & \textcolor{black}{32.82}  
		& \textcolor{black}{37.11}  & \textcolor{black}{38.33}  
		& \textcolor{black}{39.99}  & \textcolor{black}{40.22}   
		& \textcolor{black}{9.62}  & \textcolor{black}{14.81}  & \textcolor{black}{17.37}  & \textcolor{black}{18.51}  & \textcolor{black}{19.68}  & \textcolor{black}{20.22}   
		& \textcolor{black}{9.26}   & \textcolor{black}{14.17}  
		& \textcolor{black}{16.57}  & \textcolor{black}{17.60}  
		& \textcolor{black}{18.75}  & \textcolor{black}{19.41}   \\
		DPLM
		& \textcolor{blue}{24.54}  & \textcolor{blue}{36.10}  
		& \textcolor{blue}{40.52}  & \textcolor{blue}{42.73}  
		& \textcolor{blue}{47.06}  & \textcolor{blue}{47.41}
		&  \textcolor{blue}{11.51} &  \textcolor{blue}{16.39} &  \textcolor{blue}{18.21} &  \textcolor{blue}{19.47} &  \textcolor{blue}{21.39} &  \textcolor{blue}{21.77} 
		& \textcolor{blue}{11.92}  & \textcolor{blue}{15.35}  
		& \textcolor{blue}{16.88}  & 17.77  
		& \textcolor{blue}{19.85}  & \textcolor{blue}{20.59}     \\
		SHSR  
		& 19.23  & 35.56  & 38.73  & 39.50  & 41.60  & 41.43   
		& 10.14 & 16.01 & 18.05 & 18.99 & 19.60 & 19.66 
		&  9.30  & 14.33  & 17.03  & \textcolor{blue}{18.09}  & 18.22  & 18.13  \\
		GTH              
		& 17.48  & 25.81  & 28.61  & 31.62  & 33.98  & 33.96   
		& 8.59  & 13.05  & 15.95  & 17.14  & 18.73  & 19.53  
		& 8.47  & 12.78  & 15.13  & 16.
		49  & 17.88  & 18.39     \\
		{ATH\_U} 
		&\textcolor{red}{37.86} &\textcolor{red}{46.54} &\textcolor{red}{48.50} &\textcolor{red}{48.65} &\textcolor{red}{48.70} &\textcolor{red}{48.43} 
		&\textcolor{red}{17.40} &\textcolor{red}{23.81} &\textcolor{red}{24.96} &\textcolor{red}{25.76} &\textcolor{red}{24.88} &\textcolor{red}{24.97}  
		&\textcolor{red}{15.99} &\textcolor{red}{21.79} &\textcolor{red}{22.89} &\textcolor{red}{23.82} &\textcolor{red}{23.61} &\textcolor{red}{24.07}   \\
		\hline
		\hline
	\end{tabular}
	\label{Table2}
\end{table*}

\begin{table*}[!tb]
	\setlength{\abovecaptionskip}{-0cm}
	\setlength{\belowcaptionskip}{-0cm}
	\scriptsize
	\centering
	\setlength{\tabcolsep}{1.8mm}
	\caption{MAP (\%) results of different methods with varying code lengths from 16 to 128 on the HoCDR subtask.}
	\begin{tabular}{l|cccccc|cccccc|cccccc}
		\hline
		\hline
		\multirow{2}*{Method} & {16}   & {32}   & {48}   & {64}   & {96}   & {128}  
		& {16}   & {32}   & {48}   & {64}   & {96}   & {128} 
		& {16}   & {32}   & {48}   & {64}   & {96}   & {128}   \\
		\cline{2-19}
		& \multicolumn{6}{c|}{{C $\rightarrow$ A}}  & \multicolumn{6}{c|}{A $\rightarrow$ P} & \multicolumn{6}{c}{C $\rightarrow$ P}\\
		\hline	
		ITQ 
		& 6.16  & 10.03  & 12.07  & 13.16  & 14.09  & 14.89  
		& 11.85  & 17.37  & 20.11  & 21.63  & 22.90  & 23.14 
		& 10.58 & 15.46  & 18.37  & 19.54  & 20.61  & 21.31  \\
		SP              
		& \textcolor{black}{7.49}  & \textcolor{black}{14.04}  & \textcolor{black}{16.07}  & \textcolor{black}{17.09}  & \textcolor{black}{17.72}  & \textcolor{black}{17.98}  
		& \textcolor{black}{16.06}  & \textcolor{black}{24.67}  & \textcolor{black}{26.86}  & \textcolor{black}{27.79}  & \textcolor{black}{28.55}  & \textcolor{black}{28.33} 
		& \textcolor{black}{13.13}  & \textcolor{black}{19.89}  & \textcolor{black}{22.89}  & \textcolor{black}{23.94}  & \textcolor{black}{24.79}  & \textcolor{black}{24.92}  \\
		DPLM 
		& 10.53 & 15.76 & \textcolor{blue}{18.68} & \textcolor{blue}{19.94} & \textcolor{blue}{21.95} & \textcolor{blue}{22.00}
		& \textcolor{blue}{18.35}  & 26.22  & \textcolor{blue}{29.56}  & \textcolor{blue}{32.61}  & \textcolor{blue}{34.13}  & \textcolor{red}{35.48}   
		& \textcolor{blue}{17.14} & 24.51 & \textcolor{blue}{26.74} & \textcolor{blue}{28.54} & \textcolor{blue}{29.23} & \textcolor{red}{30.34}  \\
		SHSR 
		& \textcolor{blue}{13.22} & \textcolor{blue}{17.87} & \textcolor{blue}{18.68} & 18.60 & 18.35 & 16.55 
		& 17.67  & \textcolor{blue}{27.75}  & 27.17  & 28.13  & 26.24  & 24.30  
		& 15.73 & \textcolor{blue}{24.99} & 26.54 & 26.99 & 25.36 & 23.72  \\
		GTH               
		& 7.15  & 10.09  & 12.08  & 13.06  & 14.51  & 14.88 
		& 10.91  & 17.44  & 20.46  & 21.42  & 22.50  & 22.91  
		& 10.93  & 16.48  & 18.34  & 18.69  & 19.24  & 19.55   \\
		{ATH\_U} 
		&\textcolor{red}{17.00} &\textcolor{red}{23.76} &\textcolor{red}{25.68} &\textcolor{red}{27.05} &\textcolor{red}{27.21} &\textcolor{red}{26.49}  
		&\textcolor{red}{24.03} &\textcolor{red}{32.03} &\textcolor{red}{33.45} &\textcolor{red}{34.15} &\textcolor{red}{34.85} &\textcolor{blue}{34.14}  
		&\textcolor{red}{20.64} &\textcolor{red}{27.13} &\textcolor{red}{28.49} &\textcolor{red}{29.81} &\textcolor{red}{29.72} &\textcolor{blue}{29.87}     \\
		\hline
		\hline
	\end{tabular}
	\label{Table3}
\end{table*}

\begin{figure*}[!tb]
	\centering
	\subfigure[\label{}]{\includegraphics[scale=0.23,trim=5 5 50 20,clip]{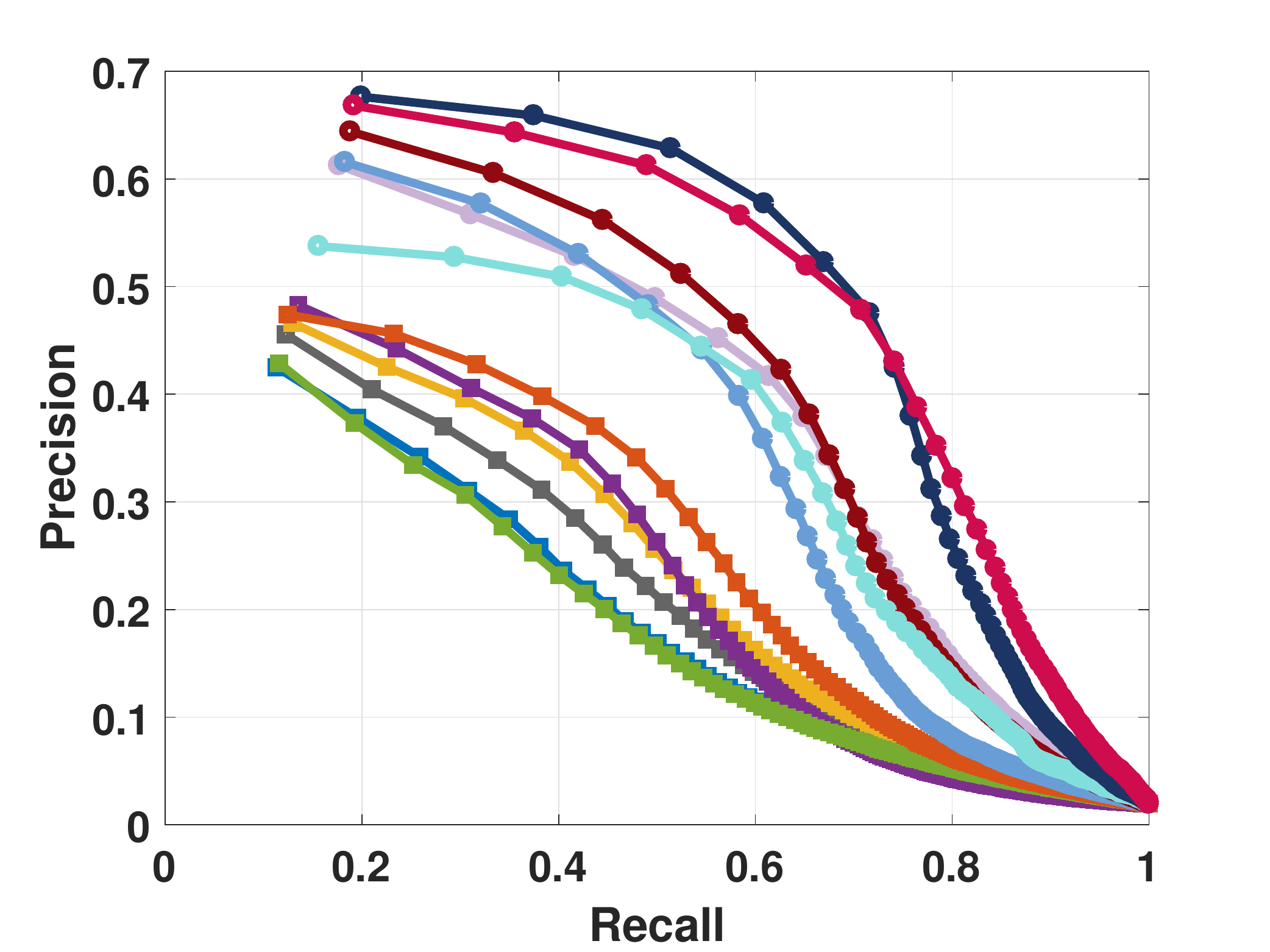}}
	\quad \subfigure[\label{}]{\includegraphics[scale=0.23,trim=5 5 50 20,clip]{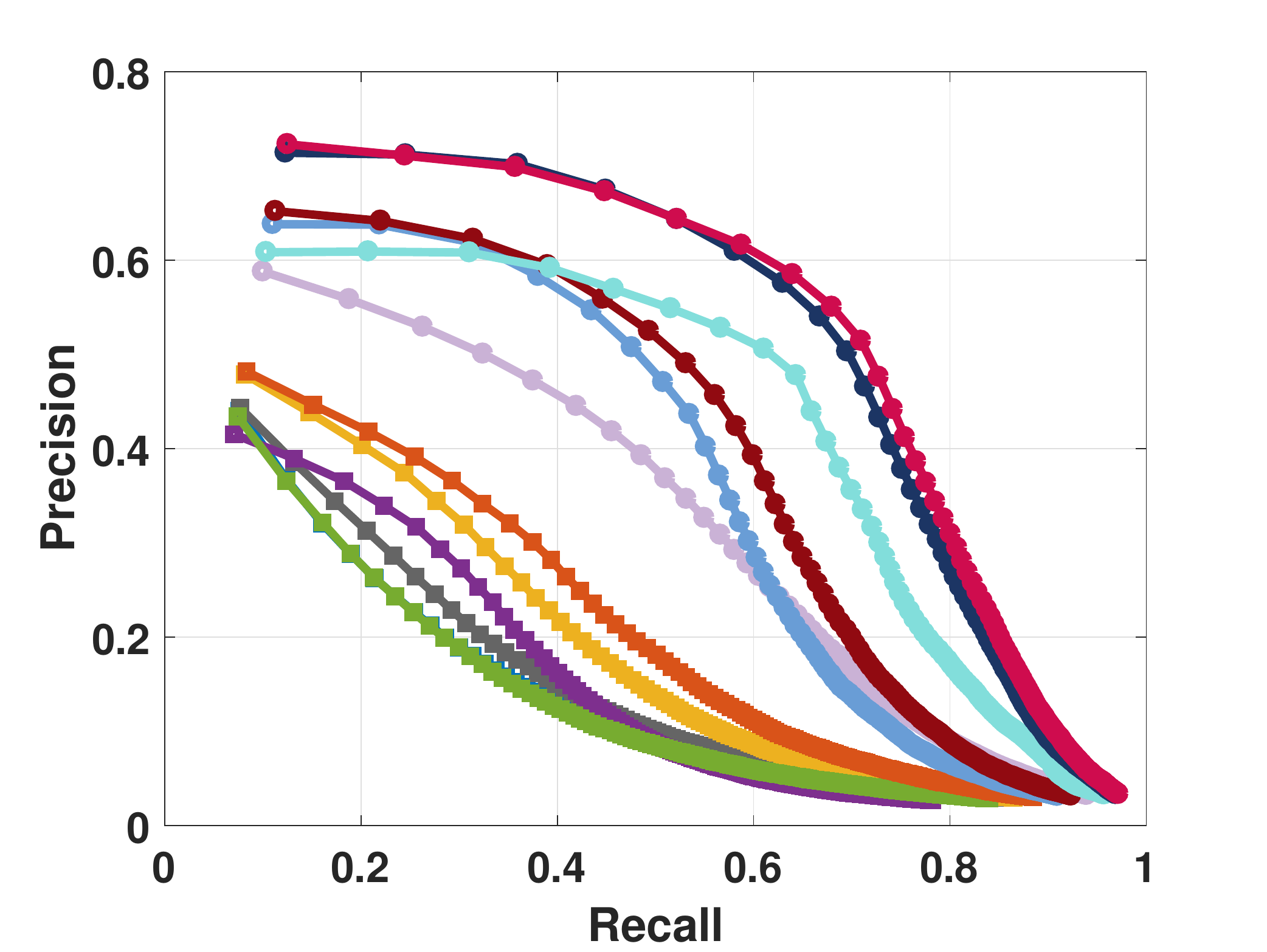}}
	\quad \subfigure[\label{}]{\includegraphics[scale=0.23,trim=5 5 50 20,clip]{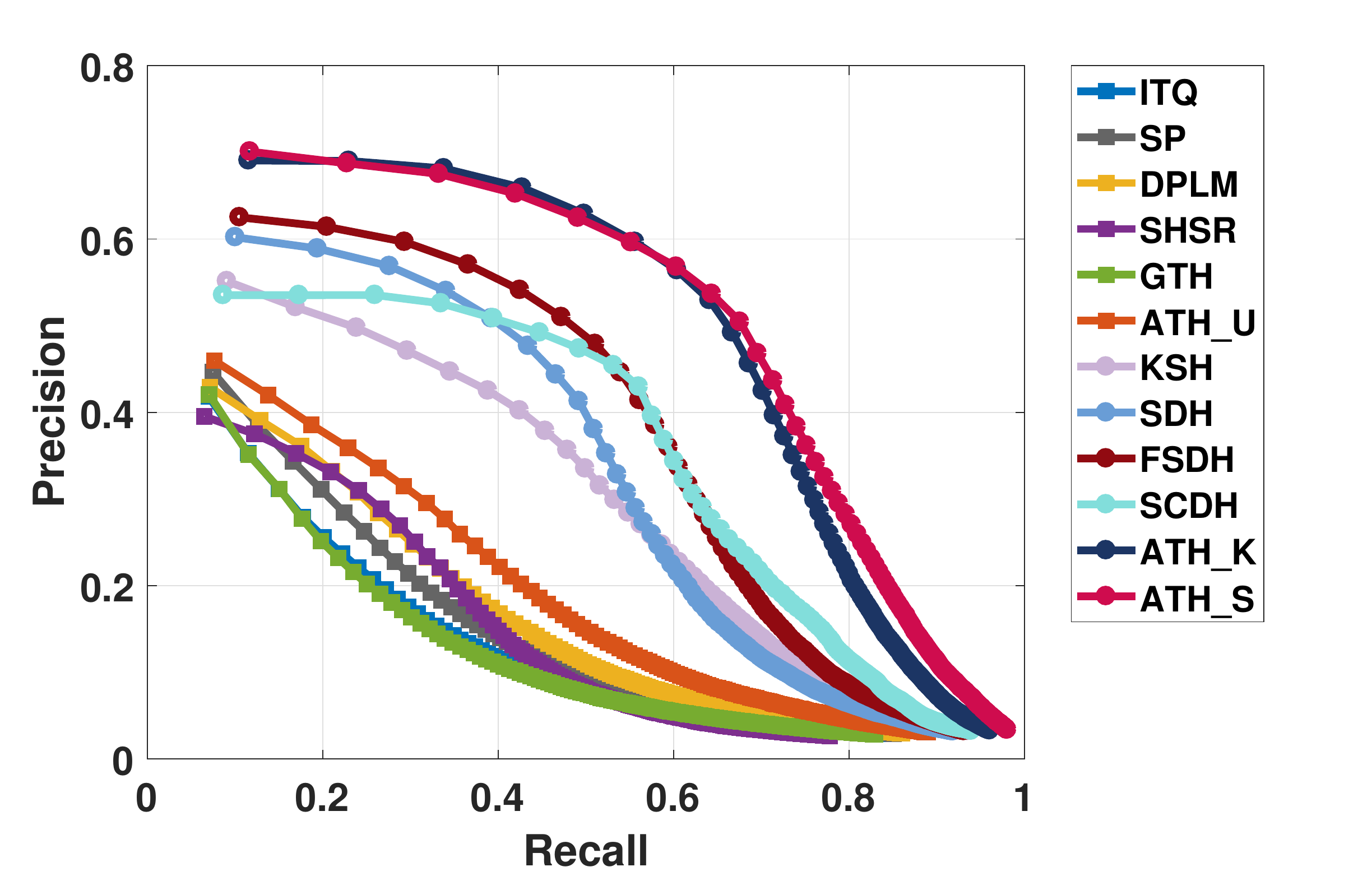}}
	\caption{PR curves on the HoSDR subtask. (a) P $\rightarrow$ A, (b) R $\rightarrow$ C, (c) P$\rightarrow$ C. }
	\label{PR1}
\end{figure*}

\begin{figure*}[!tb]
	\centering
	\subfigure[\label{}]{\includegraphics[scale=0.23,trim=5 5 50 20,clip]{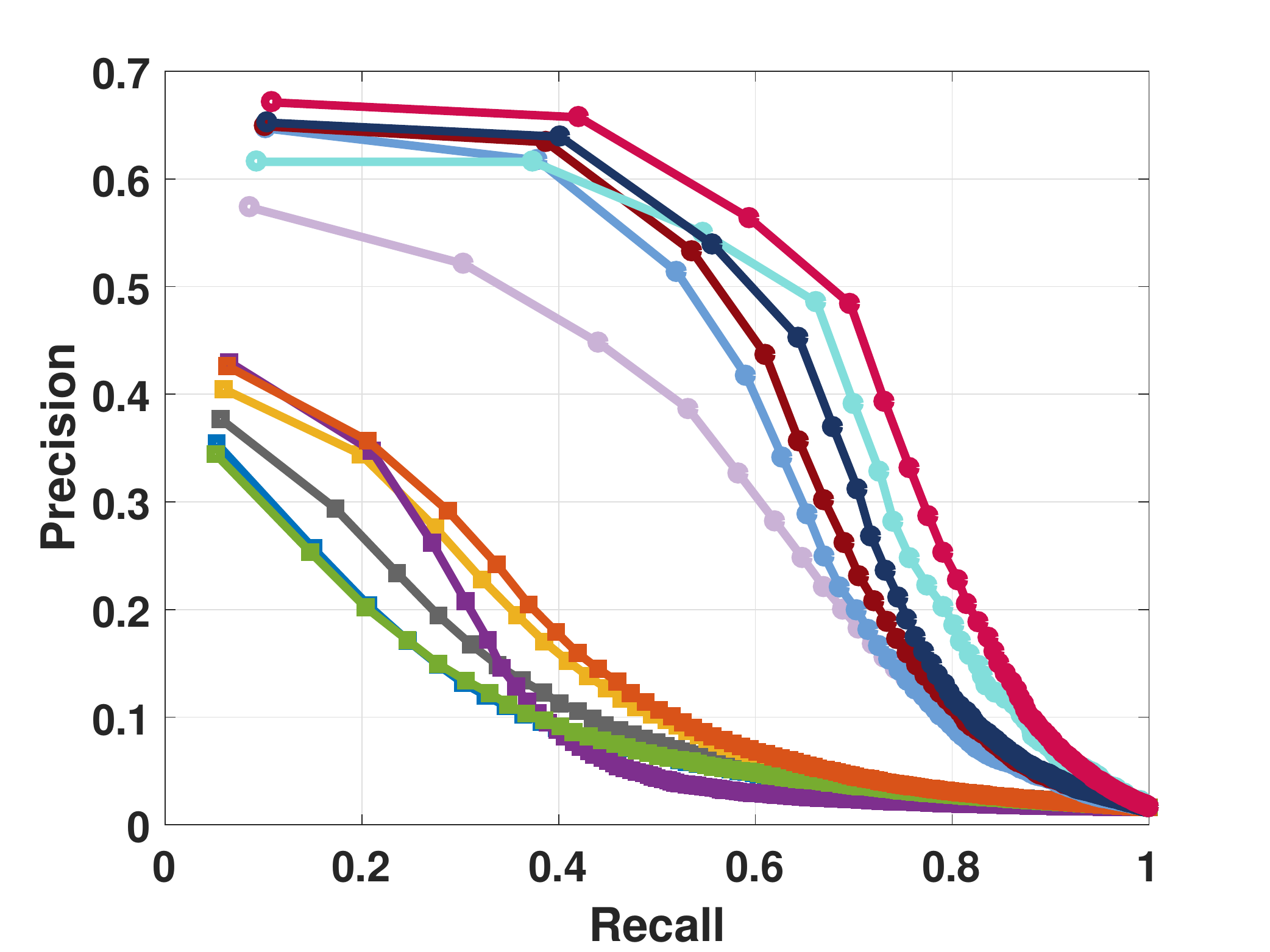}}
	\quad \subfigure[\label{}]{\includegraphics[scale=0.23,trim=5 5 50 20,clip]{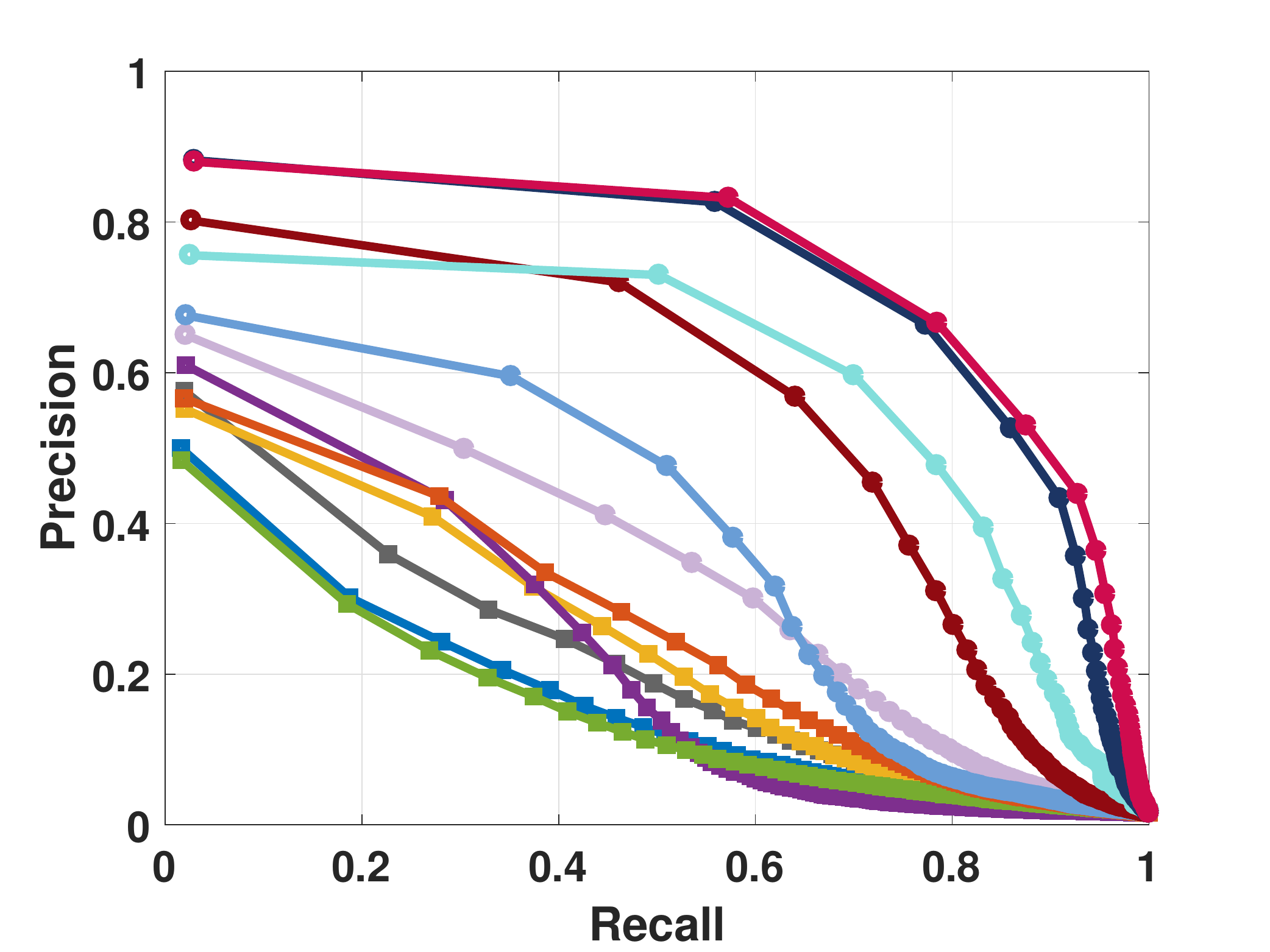}}
	\quad \subfigure[\label{}]{\includegraphics[scale=0.23,trim=5 5 50 20,clip]{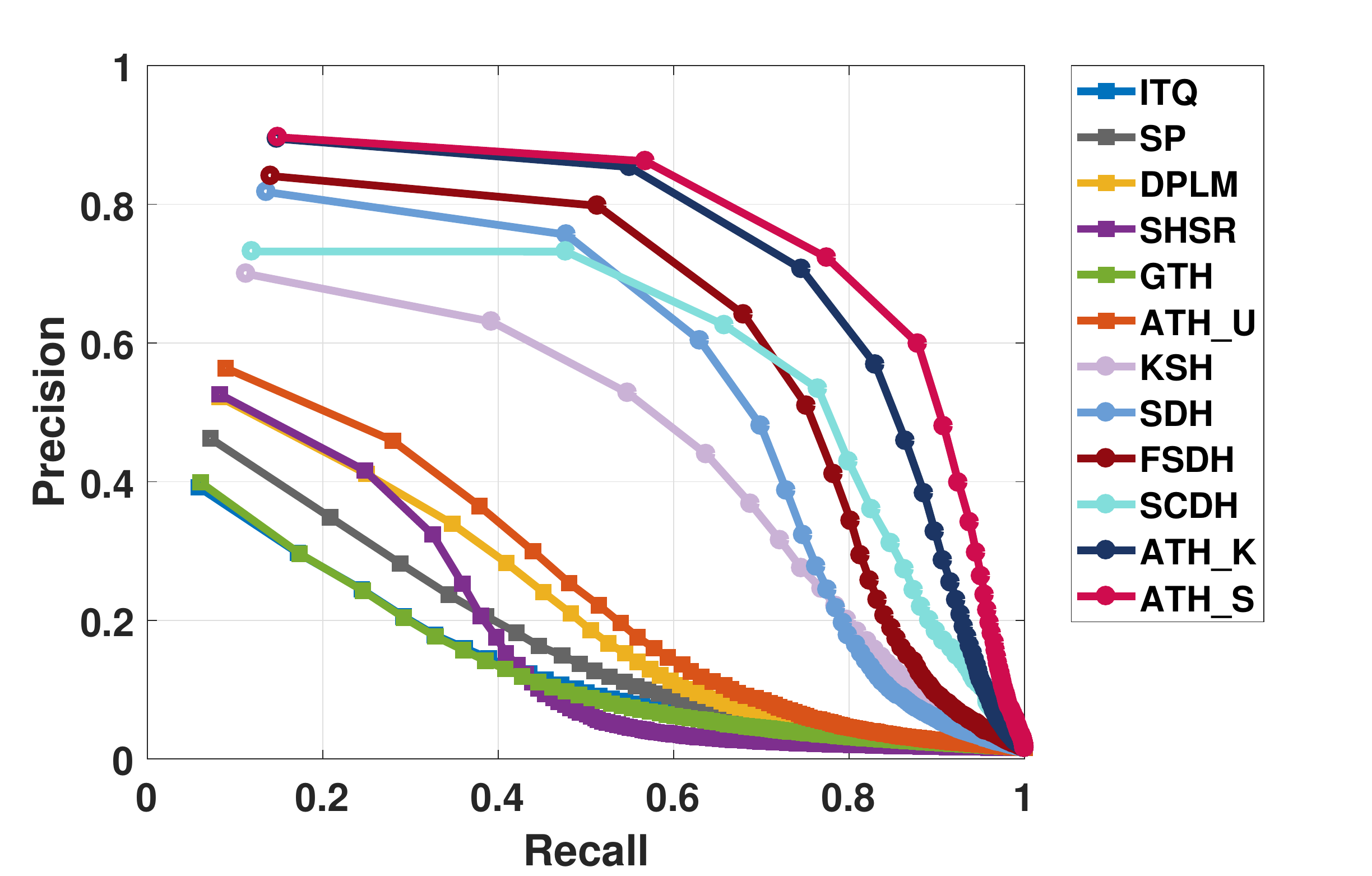}}
	\caption{PR curves on the HoCDR subtask. (a) C $\rightarrow$ A, (b) A $\rightarrow$ P, (c) C $\rightarrow$ P.} 
	\label{PR2}
\end{figure*}

\section{Experiments}
In our experiments, we compare the proposed ATH\_U with unsupervised hashing methods ITQ \cite{ITQ}, SP \cite{SP}, DPLM \cite{DPLM}, SHSR \cite{SHSR}, and transfer hashing method GTH \cite{GTH}. For ATH\_K and ATH\_S, we compare them with supervised hashing methods KSH \cite{KSH}, SDH \cite{SDH}, FSDH \cite{FSDH}, and SCDH \cite{SCDH}. For ATH\_M, we compare it with KSH, SDH and transfer hashing methods PWCF \cite{PWCF} and DHLing \cite{DHLing}.

\subsection{Datasets} 
To evaluate the performance of our ATH framework, we conduct experiments both on the single-domain and cross-domain datasets. The used datasets are described as follows:
\subsubsection{Single-domain Datasets} We first evaluate the proposed ATH\_U on single-domain datasets, including CIFAR-10\footnote{http://www.cs.toronto.edu/~kriz/cifar.html} and SUN397 \cite{SUN397} datasets. 
The CIFAR-10 dataset contains $60,000$ images of $10$
classes, where $59,000$ images ($5,900$ images sampled from each class) are randomly selected for training and the rest images are used for testing. The SUN397 dataset includes $108,753$ scene
images from $397$ well-sampled categories, where each category
contains at least $100$ samples. In our experiments, the training set contains $106,953$ images and the rest $1,800$ images are used for testing.

\subsubsection{Cross-domain Datasets}
For the proposed GITR task, we test our ATH framework on three cross-domain benchmarks, including \texttt{Office Home} \cite{OfficeHome}, \texttt{Caltech\&Amazon} \cite{GFK} and \texttt{VisDA} \cite{VisDA}. Specifically, for HoSDR and HoCDR subtasks, we follow \cite{PWCF, DHLing} and conduct experiments on \texttt{Office Home}, which is a well-known cross-domain benchmark containing four different domains: Artistic (A), Clip Art (C), Product (P), and Real-World (R). 
For simplicity, we use $\#\rightarrow\&$ to indicate that the source domain is $\#$ and the target domain is $\&$. 
For HeSDR and HeCDR subtasks, we conduct experiments on \texttt{Caltech\&Amazon} and \texttt{VisDA} benchmarks. 
\texttt{Caltech\&Amazon} is constructed from Office-Caltech10 dataset \cite{GFK}, which includes images of $10$ classes from different domains. 
\texttt{VisDA} consists of a synthetic domain and two real-image domains containing the same $12$ object categories. The training set of \texttt{VisDA} contains synthetic images generated from $3$D CAD models and the validation set includes real images selected from COCO dataset \cite{COCO}. In our experiments, we set the training set of \texttt{VisDA} as source domain and validation set as target domain. For each domain of \texttt{VisDA}, we randomly select $500$ images from each class for training. 
For unsupervised methods, all samples in the source and target domains are used for training. For semi-supervised methods, since there are no label samples in the target domain, we only use the samples in the source domain for training. Note that, the target domain samples can be used for training by our proposed ATH\_M method.

\subsection{Experimental Setup}
\subsubsection{Feature Extraction}
For CIFAR-10 dataset, each image is transformed to a $512$-dimensional GIST \cite{GIST} feature vector.  For SUN397 dataset, each image is reshaped to a $1,600$-dimensional feature vector. 
For \texttt{Office Home}, we follow \cite{PWCF, DHLing} and each image is represented by a \textit{$4096$-dimensional} feature extracted by VGG-$16$ network \cite{VGG}, thereby providing a homogeneous feature space. 
For \texttt{Caltech\&Amazon}, \texttt{Amazon} utilizes SURF features and each image is encoded by \textit{$800$-bin} histograms, while \texttt{Caltech} adopts deep features and each image is encoded by a \textit{$4096$-dimensional} CNN feature vector \cite{DeCAF}. For \texttt{VisDA}, the source images are extracted by ResNet-$50$ network \cite{ResNet} (obtaining \textit{$2048$-dimensional} features) while the target images are extracted by VGG-$16$ network \cite{VGG} (obtaining \textit{$4096$-dimensional} features).

\subsubsection{Dimensionality Reduction} For HeSDR and HeCDR subtasks, the known hashing methods cannot be directly applied since the feature dimensions of source and target domains are misaligned. For comparison, we first perform dimensionality reduction and then take the aligned features as the inputs of comparison methods.
Specifically, if the feature dimension of source domain is larger than that of target domain, PCA \cite{PCA} dimensionality reduction is performed on source domain, and the number of reduced features is determined by the feature dimension of target domain. Otherwise, we perform dimensionality reduction on target domain.
Note that, our methods do not require such a feature preprocessing, thus avoiding information loss caused by dimensionality reduction.

\begin{table*}[!tb]
	\setlength{\abovecaptionskip}{-0cm}
	\setlength{\belowcaptionskip}{-0cm}
	\scriptsize
	\centering
	\setlength{\tabcolsep}{1.8mm}
	\caption{MAP (\%) results of different methods with varying code lengths from 16 to 128 on the HeSDR subtask.}
	\begin{tabular}{l|cccccc|cccccc|cccccc}
		\hline
		\hline
		\multirow{2}*{Method} & {16}   & {32}   & {48}   & {64}   & {96}   & {128}  
		& {16}   & {32}   & {48}   & {64}   & {96}   & {128} 
		& {16}   & {32}   & {48}   & {64}   & {96}   & {128}   \\
		\cline{2-19}
		& \multicolumn{6}{c|}{\texttt{Am} ($4096$d) $\rightarrow$ \texttt{Ca} ($800$d)} 
		& \multicolumn{6}{c|}{\texttt{Ca} ($800$d) $\rightarrow$ \texttt{Am} ($4096$d)}  
		& \multicolumn{6}{c}{\texttt{VisDA} train ($2048$d) $\rightarrow$ validation ($4096$d)}\\
		\hline
		ITQ 
		&21.40 & {25.02} & {25.95} & {26.60} & {27.04} & {27.51} & 53.69 & 61.02 & 62.30 & 61.85 & 60.48 & 59.69 & 44.19 & 49.88 & 49.17 & 49.82 & 50.25 & 50.33 \\
		SP 
		&{22.01} & \textcolor{red}{27.98} & \textcolor{red}{29.15} & \textcolor{red}{28.71} & \textcolor{red}{29.16} & \textcolor{red}{29.14} & {65.84} & {72.37} & 72.64 & 70.42 & 67.67 & 65.50 & 56.71 & 62.34 & 61.25 & 60.90 & 59.79 & 59.17 \\
		DPLM 
		& 17.92 & 21.60 & 21.44 & 23.79 & 23.30 & 24.15
		& 60.78 & 69.05 & {72.81} & {75.27} & \textcolor{blue}{76.06} & \textcolor{red}{78.03} 
		& 55.67 & \textcolor{blue}{64.68} & \textcolor{red}{69.56} & \textcolor{red}{70.61} & \textcolor{red}{71.34} & \textcolor{red}{71.81}      \\
		SHSR 
		& 19.73 & 22.35 & 20.70 & 22.33 & 24.35 & 24.27 
		& \textcolor{blue}{72.99} & \textcolor{red}{77.39} & \textcolor{red}{77.00} & \textcolor{red}{76.82} & {75.44} & \textcolor{blue}{75.59}
		& \textcolor{blue}{64.06} & \textcolor{red}{67.00} & 59.62 & 56.04 & 53.25 & 50.60       \\
		GTH 
		&\textcolor{blue}{22.28} & 23.44 & 25.27 & 26.00 & 26.52 & 26.93 
		& 51.52 & 52.36 & 53.68 & 54.61 & 55.19 & 57.11
		& 39.71 & 44.86 & 50.20 & 51.03 & 52.65 & 53.23\\
		{ATH\_U}   
		& \textcolor{red}{23.87} & \textcolor{blue}{25.67} 
		& \textcolor{blue}{27.42} & \textcolor{blue}{28.36} 
		& \textcolor{blue}{28.30} & \textcolor{blue}{28.22} 
		& \textcolor{red}{74.84} & \textcolor{blue}{76.36} & \textcolor{blue}{75.17} & \textcolor{blue}{75.64} & \textcolor{red}{76.43} & 75.54 
		& \textcolor{red}{66.13} & 63.42 & \textcolor{blue}{65.98} & \textcolor{blue}{69.29} & \textcolor{blue}{69.01} & \textcolor{blue}{69.45} \\              
		\hline
		\hline
	\end{tabular}
	\label{Table4}
\end{table*}

\begin{table*}[!tb]
	\setlength{\abovecaptionskip}{-0cm}
	\setlength{\belowcaptionskip}{-0cm}
	\scriptsize
	\centering
	\setlength{\tabcolsep}{1.8mm}
	\caption{MAP (\%) results of different methods with varying code lengths from 16 to 128 on the HeCDR subtask.}
	\begin{tabular}{l|cccccc|cccccc|cccccc}
		\hline
		\hline
		\multirow{2}*{Method} & {16}   & {32}   & {48}   & {64}   & {96}   & {128}  
		& {16}   & {32}   & {48}   & {64}   & {96}   & {128} 
		& {16}   & {32}   & {48}   & {64}   & {96}   & {128}   \\
		\cline{2-19}
		& \multicolumn{6}{c|}{\texttt{Am} ($4096$d) $\rightarrow$ \texttt{Ca} ($800$d)} 
		& \multicolumn{6}{c|}{\texttt{Ca} ($800$d) $\rightarrow$ \texttt{Am} ($4096$d)}  
		& \multicolumn{6}{c}{\texttt{VisDA} train ($2048$d) $\rightarrow$ validation ($4096$d)} \\
		\hline
		ITQ 
		& 9.76  & \textcolor{blue}{10.10} & 9.86  & 9.92  & 9.78  & 9.79  & 9.24  & 9.72  & 9.71  & 9.78  & \textcolor{blue}{9.80}  & 9.83  & 8.23  & 7.90 & 8.10 & 7.99  & 8.25  & 8.31  \\
		SP   
		& \textcolor{blue}{9.93}  & 9.72  & 9.64  & 9.60  & 9.79  & 9.78  & 9.21  & 9.60  & 9.66  & 9.61  & 9.63  & 9.73  & 8.12  & 8.14 & 8.05 & 7.94  & 7.86  & 7.95  \\
		DPLM 
		&  9.61 &  9.96 &  \textcolor{blue}{9.91} &  9.61 & \textcolor{blue}{10.14} &  9.88
		& \textcolor{blue}{10.30} & \textcolor{blue}{10.48} & \textcolor{blue}{10.37} & 10.15 &  9.78 &  \textcolor{blue}{9.98}
		&  8.59 &  \textcolor{blue}{8.90} &  8.49 &  8.14 &  \textcolor{blue}{9.08} &  7.94      \\
		SHSR 
		&  9.76 &  9.96 &  9.62 &  \textcolor{blue}{9.98} &  9.91 &  \textcolor{blue}{9.91} 
		&  9.93 &  9.86 & 10.12 & \textcolor{blue}{10.33} &  9.74 &  9.93
		&  \textcolor{blue}{8.89} &  8.13 &  \textcolor{blue}{8.54} &  \textcolor{blue}{8.39} &  8.24 &  \textcolor{blue}{8.47}      \\
		GTH 
		& 9.70  & 9.65  & 9.71  & 9.74  & 9.67  & 9.61  & 9.45  & 9.33  & 9.56  & 9.40  & 9.57  & 9.61  & 7.72  & 7.92 & 8.08 & 8.09  & 8.16  & 8.02  \\
		{ATH\_U}
		& \textcolor{red}{12.21} & \textcolor{red}{13.14} 
		& \textcolor{red}{13.84} & \textcolor{red}{13.85} 
		& \textcolor{red}{13.02} & \textcolor{red}{12.08} 
		& \textcolor{red}{15.45} & \textcolor{red}{15.77} 
		& \textcolor{red}{18.57} & \textcolor{red}{18.01} 
		& \textcolor{red}{20.45} & \textcolor{red}{15.92} 	
		& \textcolor{red}{12.58} & \textcolor{red}{11.22}
		& \textcolor{red}{12.16}  & \textcolor{red}{15.89} 
		& \textcolor{red}{13.46} & \textcolor{red}{13.46} \\
		\hline
		\hline
	\end{tabular}
	\label{Table5}
\end{table*}

\begin{figure*}[!tb]
	\centering
	\subfigure[\label{}]{\includegraphics[scale=0.23,trim=5 5 50 20,clip]{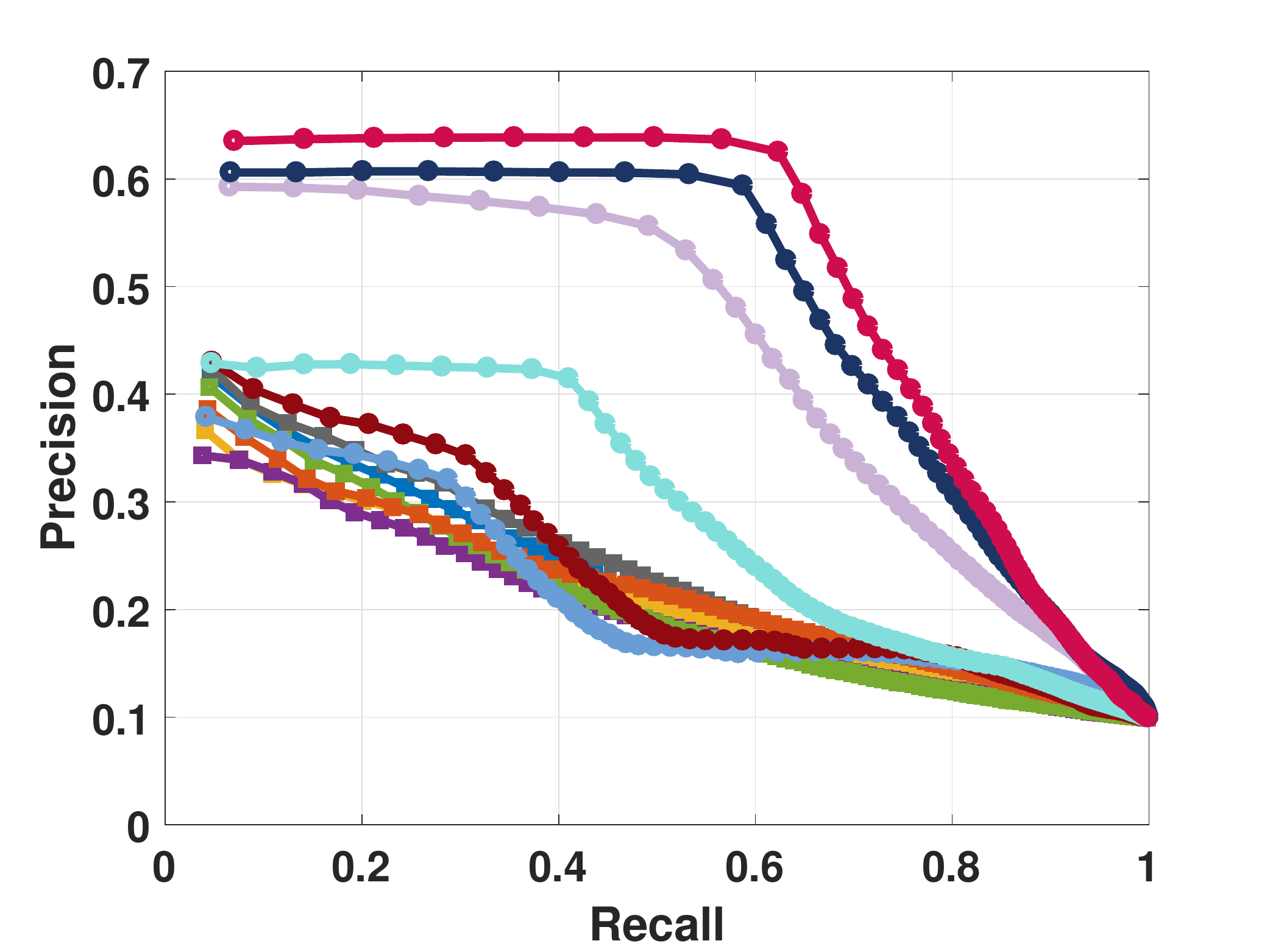}}
	\quad \subfigure[\label{}]{\includegraphics[scale=0.23,trim=5 5 50 20,clip]{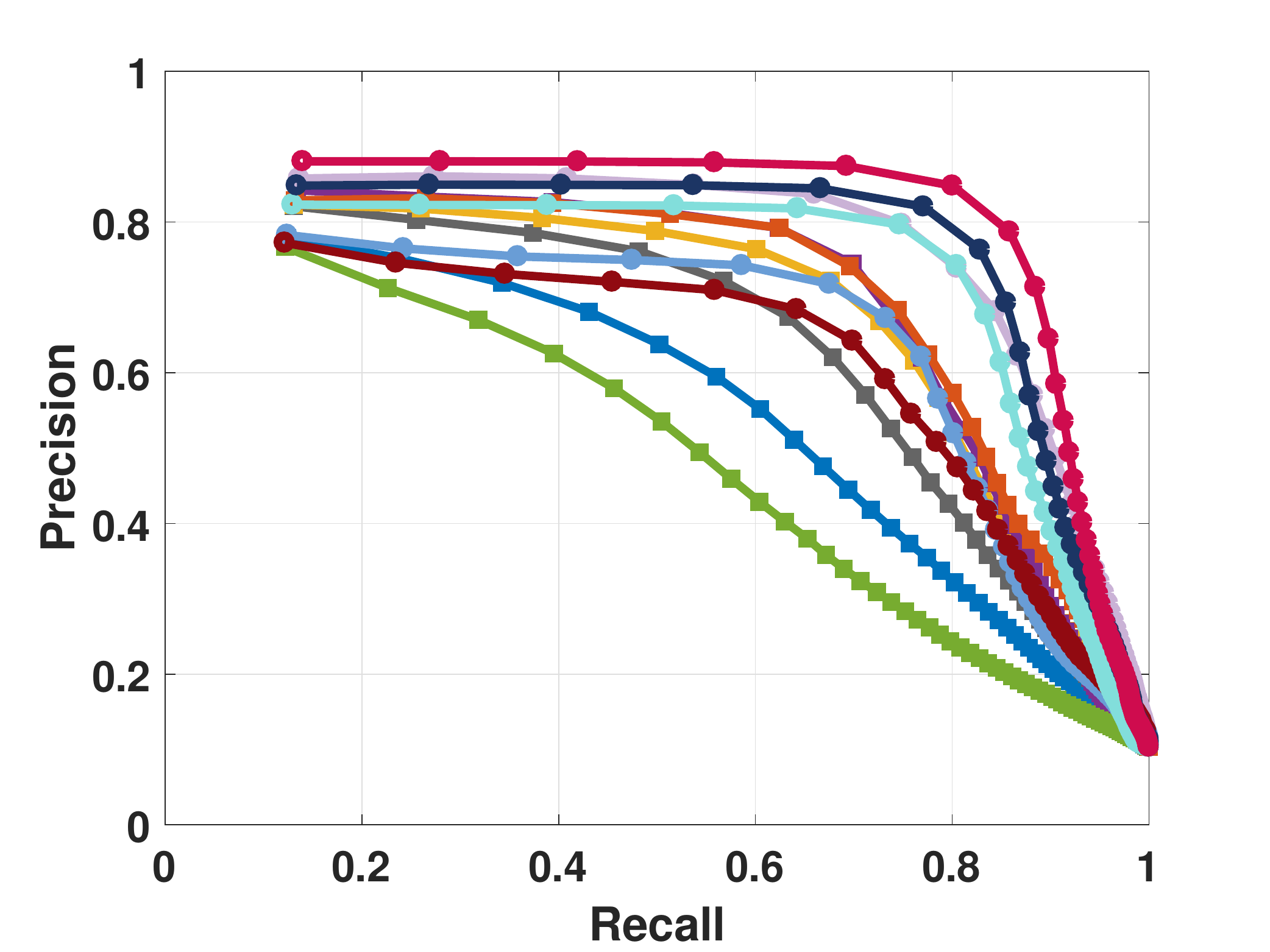}}
	\quad \subfigure[\label{}]{\includegraphics[scale=0.23,trim=5 5 50 20,clip]{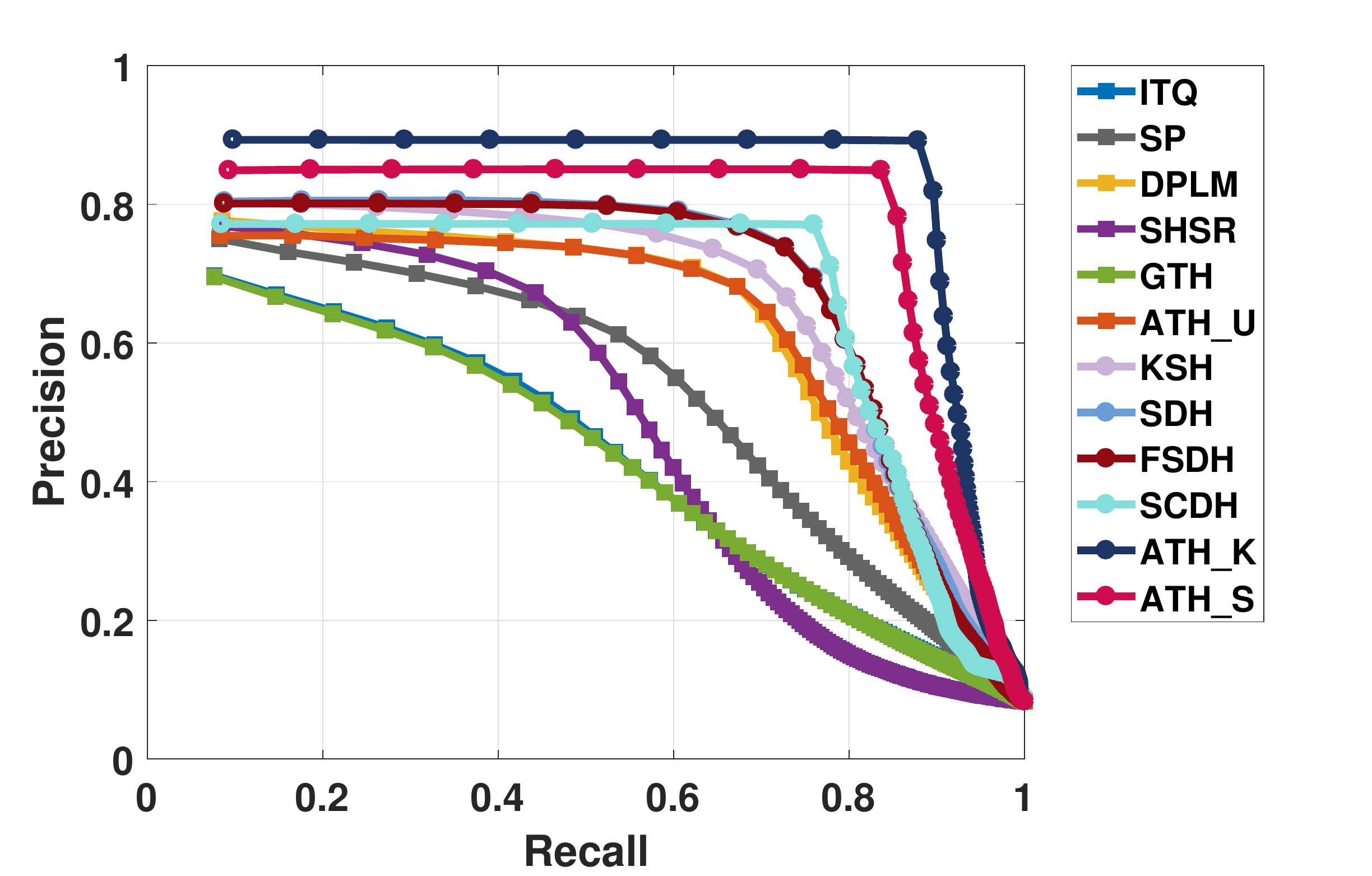}}
	\caption{PR curves on the HeSDR subtask. (a) \texttt{Amazon} $\rightarrow$ \texttt{Caltech}, (b) \texttt{Caltech} $\rightarrow$ \texttt{Amazon}, (c) \texttt{VisDa} train $\rightarrow$ validation. }
	\label{PR3}
\end{figure*}

\begin{figure*}[!tb]
	\centering
	\subfigure[\label{}]{\includegraphics[scale=0.23,trim=5 5 50 20,clip]{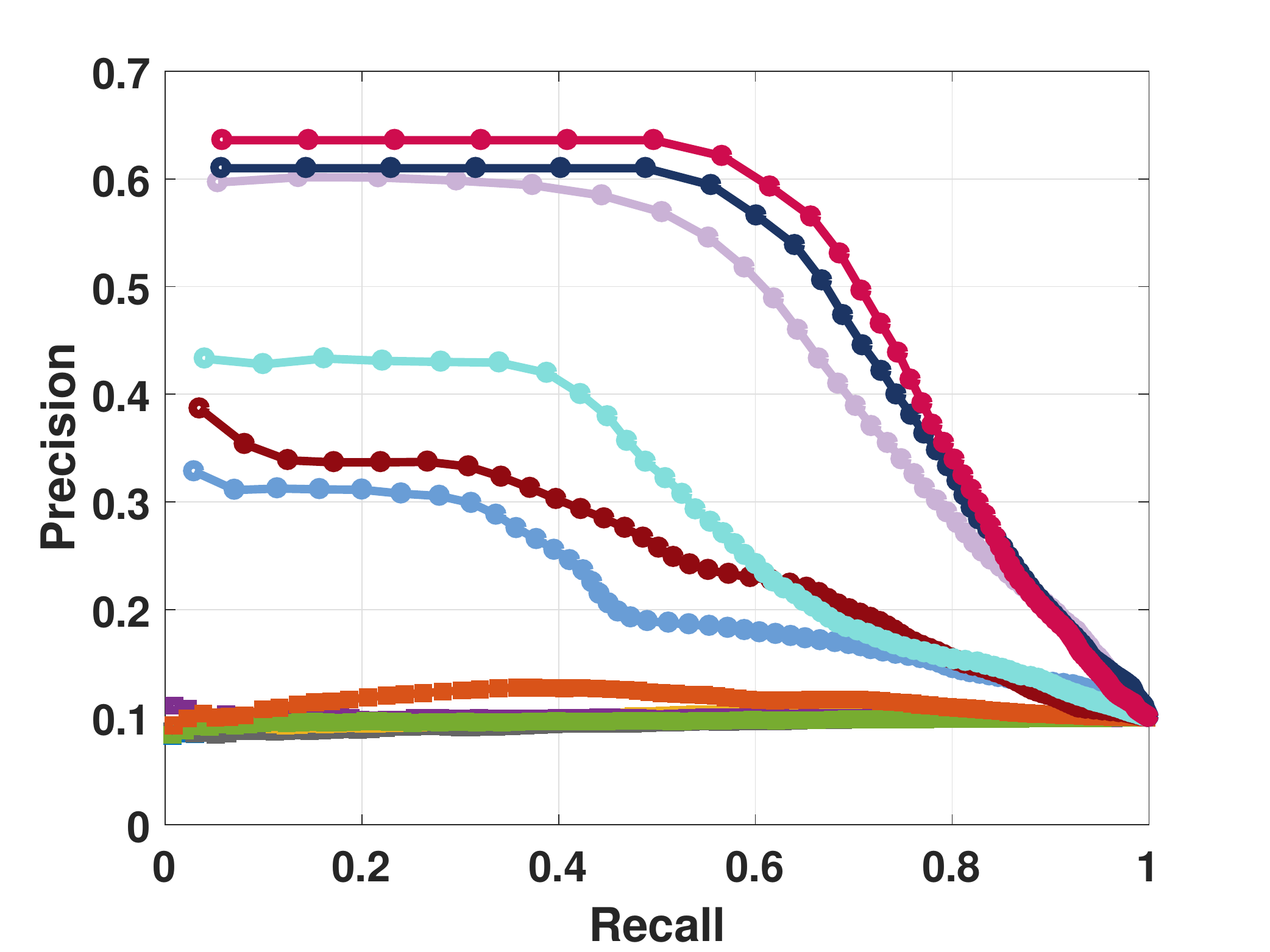}}
	\quad \subfigure[\label{}]{\includegraphics[scale=0.23,trim=5 5 50 20,clip]{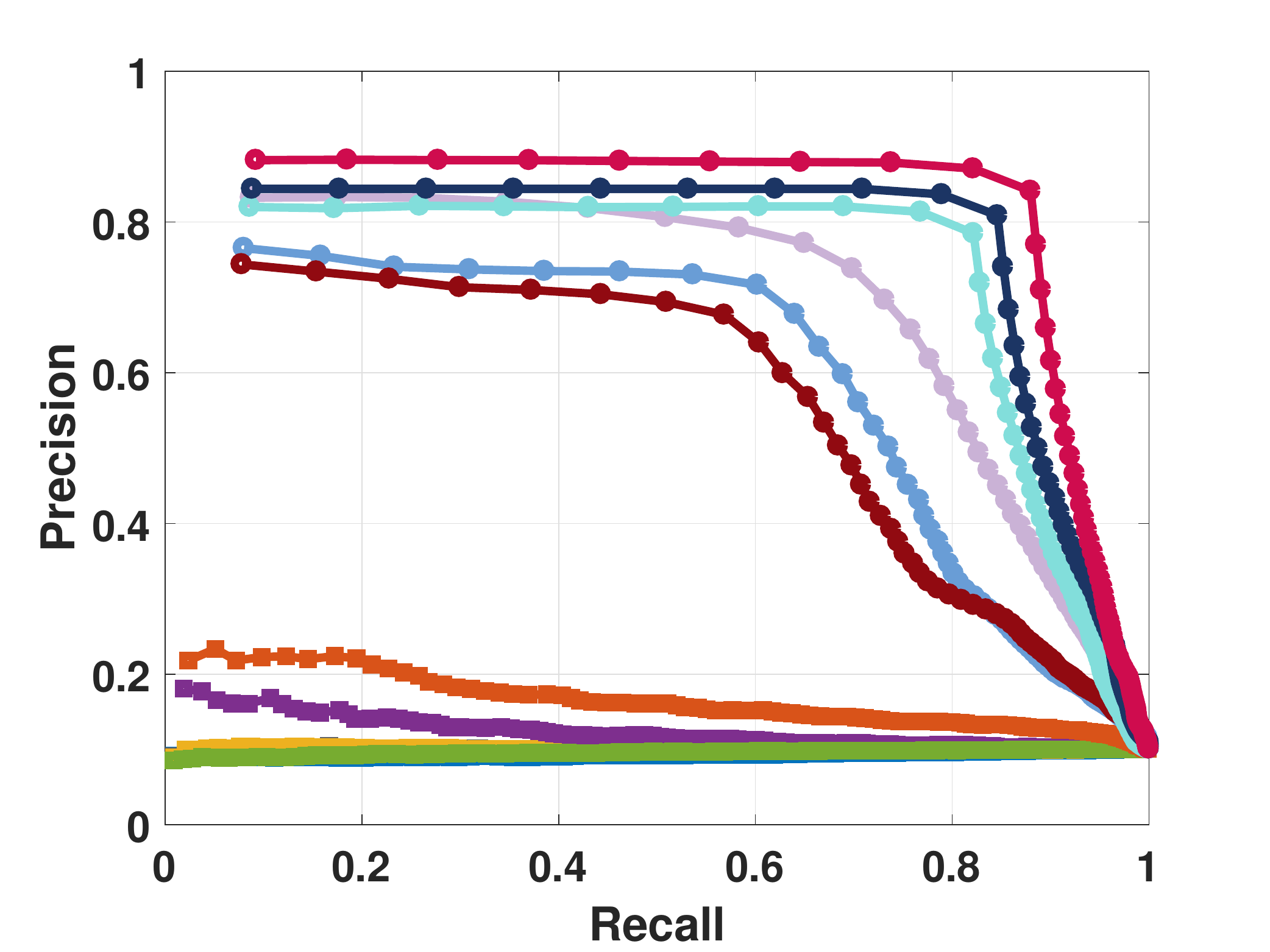}}
	\quad \subfigure[\label{}]{\includegraphics[scale=0.23,trim=5 5 50 20,clip]{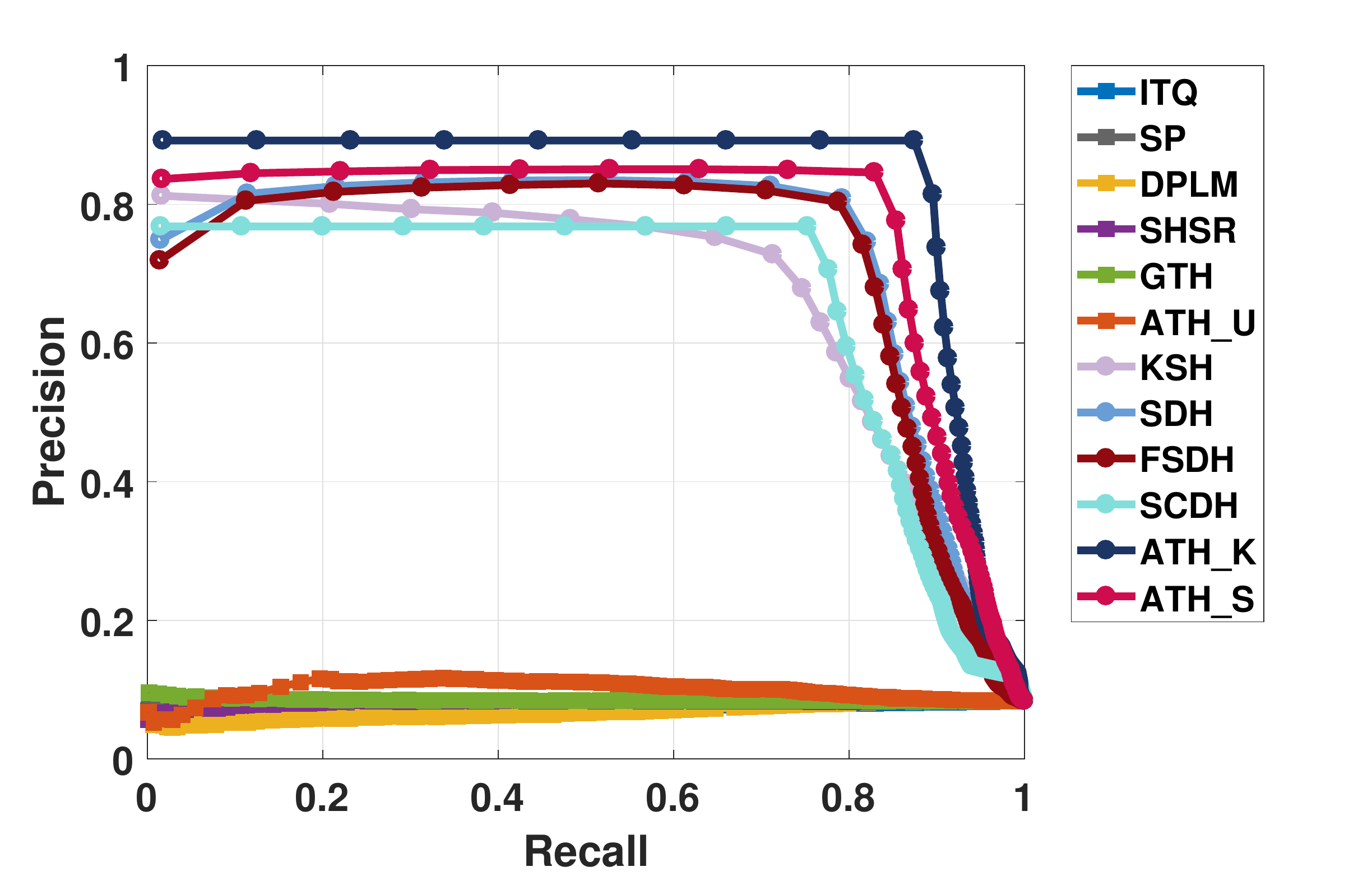}}
	\caption{PR curves on the HeCDR subtask. (a) \texttt{Amazon} $\rightarrow$ \texttt{Caltech}, (b) \texttt{Caltech} $\rightarrow$ \texttt{Amazon}, (c) \texttt{VisDa} train $\rightarrow$ validation. }
	\label{PR4}
\end{figure*}

\subsubsection{Implementation Details} The proposed ATH is implemented in MATLAB on a standard PC with Intel $2.20$-GHz CPU and $128$-GB RAM. For the comparison methods, we use their public codes and the parameter selection observes from the corresponding papers. Following the experimental settings in \cite{PWCF, DHLing}, for each dataset, we pick up $500$ samples from the target domain randomly as the query samples, and the rest images are used for training. The length of hash codes is selected from the range of $\{16, 32, 48, 64, 96, 128\}$. 
For each method, we simply set the number of nearest neighbors $\eta_k \in\{\eta_s, \eta_t\}$ to $10$ and the number of iterations $Ite$ is set to $10$.
For the parameters $\alpha_s$, $\alpha_t$, $\beta_s$, $\beta_t$, $\lambda$, we set them to $10^{-2}$, $10^{-1}$, $10^{-3}$, $10^{-1}$, $1$ for ATH\_U, $10^{-1}$, $10^{-1}$, $10^{3}$, $10^{-2}$, $1$ for ATH\_M, $10^{-2}$, $1$, $10^{-3}$, $10^{-3}$, $10^{-2}$ for ATH\_K, and $10^{-3}$, $10^{-1}$, $10^{-2}$, $10^{-2}$, $10^{-2}$ for ATH\_S, respectively.
To obtain stable results, we run $10$ independent experiments for each method and the average value of $10$ experiments is reported.

\subsubsection{Evaluation Protocol} 
Following the settings in \cite{PWCF}, we adopt the widely-used Mean Average Precision (MAP) as the evaluation metric and provide the precision-recall (PR) curves to evaluate the performance of different methods. Besides, we also provide feature visualizations to illustrate the performance differences between different methods.

\subsection{Performance on single-domain datasets} 
Table \ref{Table1} provides the MAP results of unsupervised hashing methods on the single-domain benchmarks (the highest
and second highest are marked in red and blue respectively). 
As we can see, the similarity-preservation based methods (i.e., DPLM, SHSR, ATH\_U) always achieve better performance than the projection-based (i.e., ITQ, SP, GTH) methods in the single-domain setting, which stresses the importance to preserve the similarity relationship between pairwise samples. Moreover, although the proposed ATH\_U is tailored for transfer retrieval, it can also outperform the traditional hashing methods on the single-domain benchmarks. The potential reason is that the proposed ATH\_U can adaptively learn the similarity relationship between samples from the data during the training stage.

\begin{table*}[!tb]
	\setlength{\abovecaptionskip}{-0cm}
	\setlength{\belowcaptionskip}{-0cm}
	\centering
	\setlength{\tabcolsep}{1.0mm}
	\caption{MAP (\%) results of supervised methods with 64bits on different GITR subtasks.}
	\begin{tabular}{p{1.0cm}<{\centering}|p{1.1cm}<{\centering}p{1.1cm}<{\centering}p{1.1cm}<{\centering}|p{1.2cm}<{\centering}p{1cm}<{\centering}p{1.2cm}<{\centering}|p{1.2cm}<{\centering}p{1.2cm}<{\centering}p{1.2cm}<{\centering}|p{1.2cm}<{\centering}p{1.2cm}<{\centering}p{1.2cm}<{\centering}}
		\hline	
		\hline	
		\multirow{2}*{Method}
		&\multicolumn{3}{c|}{HoSDR} &\multicolumn{3}{c|}{HoCDR} &\multicolumn{3}{c|}{HeSDR} &\multicolumn{3}{c}{HeCDR}\\
		\cline{2-13}	
		&P$\rightarrow$A  &R$\rightarrow$C  &P$\rightarrow$C  &C$\rightarrow$A  &A$\rightarrow$P  &C$\rightarrow$P &\texttt{Am} $\rightarrow$ \texttt{Ca} &\texttt{Ca} $\rightarrow$ \texttt{Am}	&\texttt{VisDA} &\texttt{Am} $\rightarrow$ \texttt{Ca} &\texttt{Ca} $\rightarrow$ \texttt{Am}	&\texttt{VisDA}  \\
		\hline	
		KSH 
		&64.33 &44.83 &41.63 &49.59 &51.78 &58.59 &53.46 &87.77 &77.38 &60.50 &76.64 &79.79 \\
		SDH  
		&62.45 &50.81 &47.32 &55.04 &44.20 &66.33 &27.10 &82.39 &79.12 &34.68 &69.20 &86.05 \\
		FSDH  
		&66.61 &55.66 &51.83 &59.14 &73.27 &70.85 &32.75 &77.44 &78.54 &41.55 &64.39 &84.75 \\
		SCDH  
		&67.23 &65.23 &57.37 &\textcolor{red}{67.73} &81.36 &81.54 &45.42 &89.45 &82.42 &45.81 &88.13 &82.51 \\
		ATH\_K  
		&\textcolor{blue}{77.33} &\textcolor{blue}{68.46} &\textcolor{blue}{64.76} &62.10 &\textcolor{blue}{90.23} &\textcolor{blue}{83.10} &\textcolor{blue}{66.87} &\textcolor{blue}{92.09} &\textcolor{blue}{89.40} &\textcolor{blue}{67.51} &\textcolor{blue}{91.62} &\textcolor{blue}{89.47} \\	
		ATH\_S 
		&\textcolor{red}{77.40} &\textcolor{red}{69.08} &\textcolor{red}{67.72} &\textcolor{blue}{65.99} &\textcolor{red}{91.32} &\textcolor{red}{86.65} &\textcolor{red}{69.08} &\textcolor{red}{93.04} &\textcolor{red}{90.15} &\textcolor{red}{70.00} &\textcolor{red}{92.52} &\textcolor{red}{89.88} \\
		\hline
		\hline
	\end{tabular}
	\label{Table6}
	\vspace{-3mm}
\end{table*}

\subsection{Performance on cross-domain datasets}
Tables \ref{Table2}-\ref{Table5} list the MAP results of unsupervised methods under four different GITR subtasks. 
Figures \ref{PR1}-\ref{PR4} provide the PR curves (with 64 bits) of all unsupervised and supervised methods under different GITR subtasks. 
As we can see, the proposed ATH\_U obtains the best performance under different settings (i.e., HoSDR, HoCDR, HeSDR and HeCDR) in most cases, which demonstrates the effectiveness and versatility of our ATH framework for different GITR subtasks.  
Moreover, we observe that the compared unsupervised methods are as effective as the proposed ATH\_U on HeSDR subtask, but their performances are worse than that of ATH\_U on HeCDR subtask. This is because the features and domains of query and retrieval sets in HeCDR subtask are misaligned and different. By simultaneously learning asymmetric hash functions and affinity relationship of cross-domain data, our method can perform knowledge transfer between domains and avoid information loss caused by feature alignment.

As we can see from Table IV, some results of DPLM are better than those of ATH\_U in the HeSDR task. The potential reason is that DPLM adopts a discrete proximal linearized minimization approach to handle the discrete constraints during the learning process \cite{DPLM}, which makes the learned binary codes more effective. Since discreteness is not the main focus of this manuscript, the proposed ATH\_U obtains binary codes by simply utilizing a relaxation strategy (i.e., the real-valued embedding are first learned, which are then thresholded to be binary codes), which may cause quantization error to some extent. However, as can be seen from Table V, when cross-domain information interaction rather than discreteness becomes the core issue in the HeCDR task, the proposed ATH\_U achieves better performance than DPLM.

\begin{table}[!tb]
	\setlength{\abovecaptionskip}{-0cm}
	\setlength{\belowcaptionskip}{-0cm}
	\centering
	\setlength{\tabcolsep}{2.5mm}
	\caption{MAP (\%) results of semi-supervised methods with 64bits on the HoCDR subtask.}
	\begin{tabular}{l|cccccc}
		\hline
		\hline
		Method&P$\rightarrow$R &R$\rightarrow$P &C$\rightarrow$R &R$\rightarrow$C &A$\rightarrow$R &R$\rightarrow$A  \\
		\cline{1-7}		
		KSH 
		& 32.02 & 34.42 & 21.56 & 18.51 & 25.87 & 20.04  \\
		SDH  
		& 25.75 & 27.90 & 15.97 & 16.72 & 32.06 & 22.79   \\
		PWCF 
		& 34.03 & 34.44 & 24.22 & 18.42 & 34.57 & 28.95 \\
		DHLing 
		& \textcolor{blue}{48.47} & \textcolor{blue}{45.24} & \textcolor{blue}{30.81} & \textcolor{blue}{25.15} & \textcolor{blue}{43.30} & \textcolor{blue}{38.68}  \\
		ATH\_M  & \textcolor{red}{50.69} & \textcolor{red}{52.16} & \textcolor{red}{33.79} & \textcolor{red}{29.14} & \textcolor{red}{50.97} & \textcolor{red}{42.80}    \\
		\hline
		\hline
	\end{tabular}
	\label{Table7}
\end{table}

 \begin{figure}[!tb]
 	\centering
 	\subfigure[Office Home]{\includegraphics[scale=0.13,trim=58 10 10 20,clip]{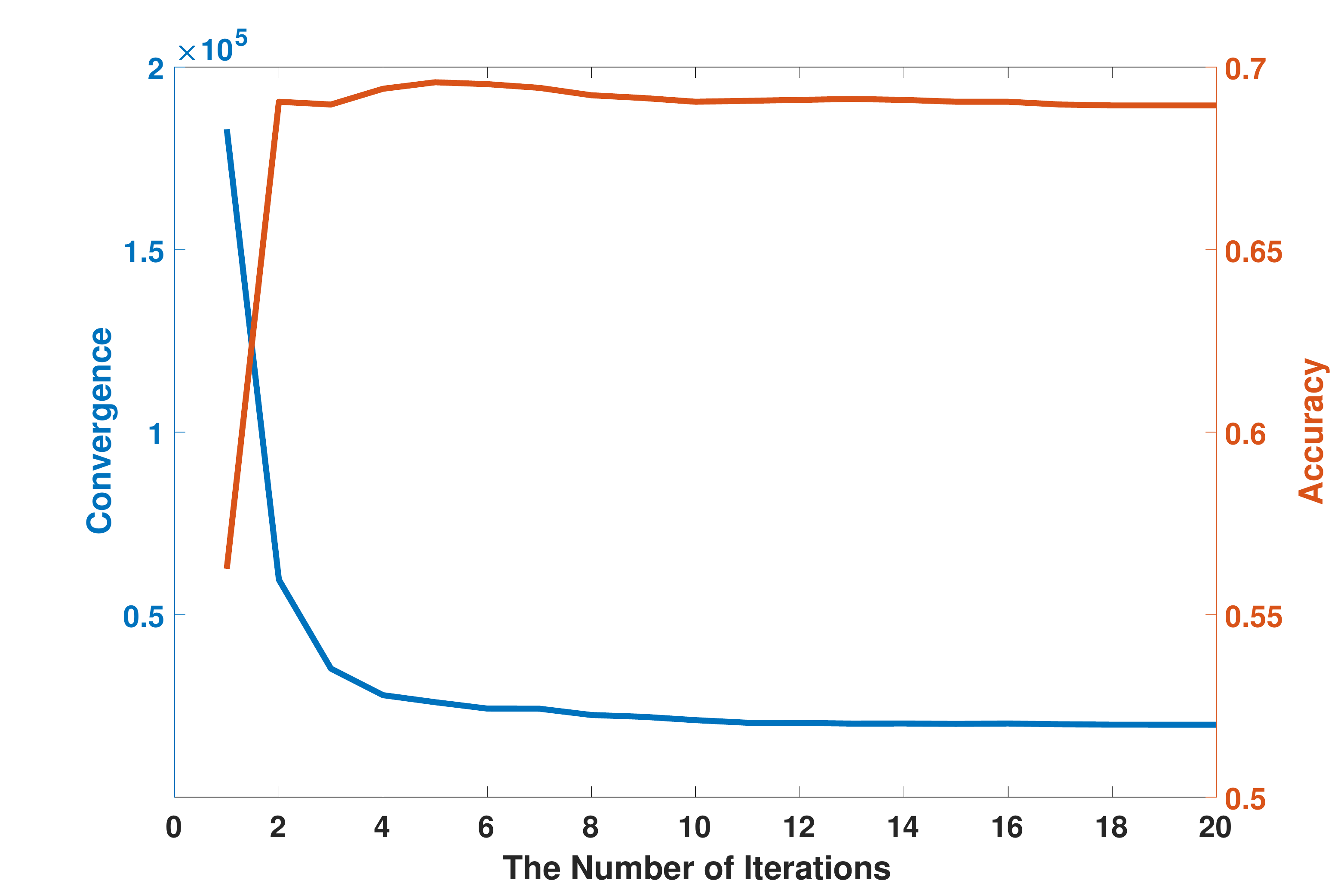}}
 	\subfigure[VisDA]{\includegraphics[scale=0.13,trim=58 10 10 20,clip]{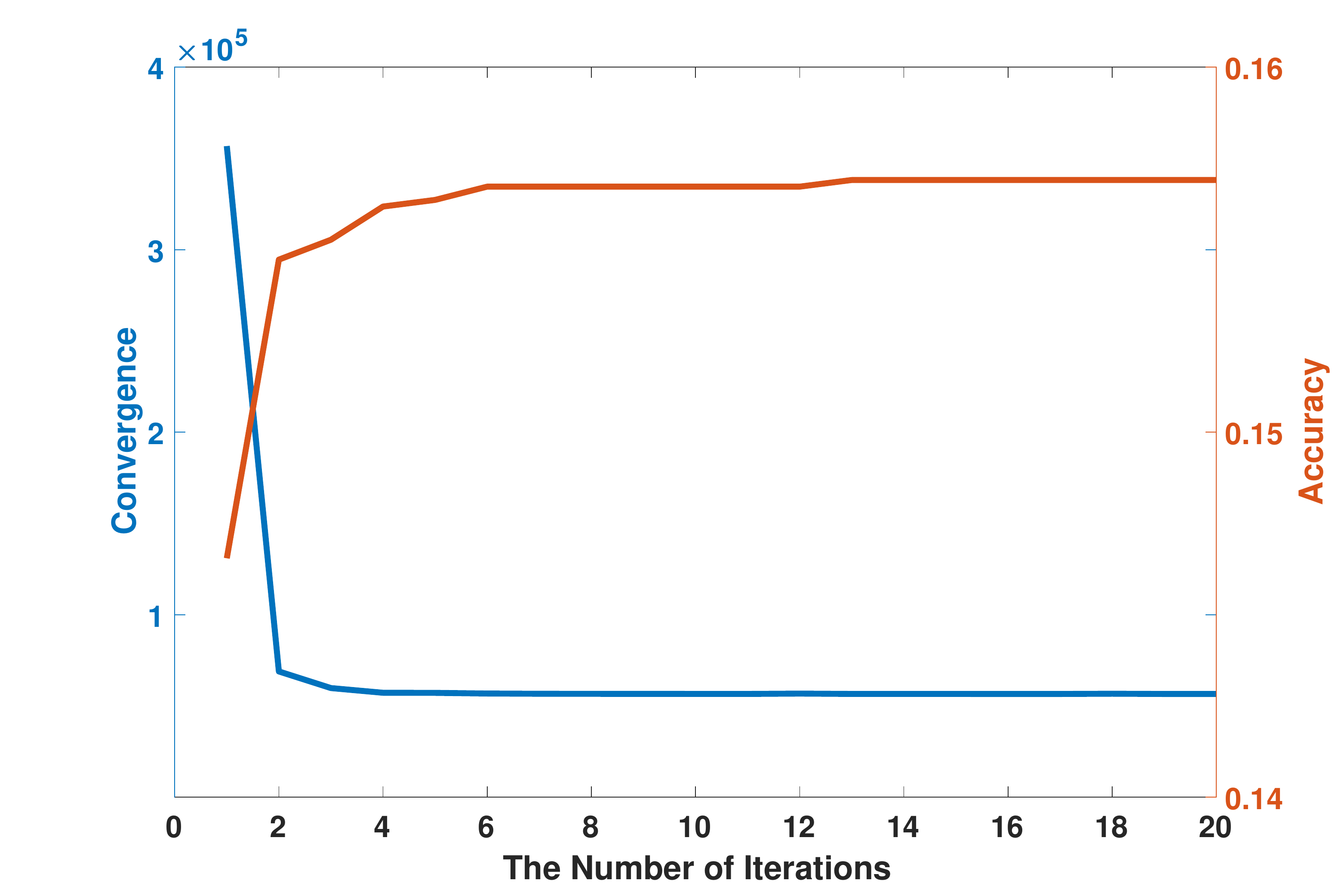}}
 	\caption{Accuracy and convergence curves on different datasets.} 
 	\label{positive}
 	\vspace{-3mm}
 \end{figure}

In addition, Table \ref{Table6} lists the MAP results of supervised methods under different GITR subtasks and \ref{Table7} provides the MAP results of different methods in the semi-supervised setting. Since the source code of DHLing is not publicly available, we directly cite the results of DHLing with 64bits using the same experimental settings with ours. From the experimental results, we observe that our ATH\_M, ATH\_S and ATH\_K method obtain better retrieval performance than the state-of-the-art hashing methods. The reason is that our methods can perform positive transfer during training, as illustrated in the next section.

\begin{table}[!tb]
	\setlength{\abovecaptionskip}{-0cm}
	\setlength{\belowcaptionskip}{-0cm}
	\centering
	\setlength{\tabcolsep}{3.5mm}
	\caption{MAP (\%) results of variants of ATH\_U (with 64 bits).}
	\begin{tabular}{c|c|c|c|c}
		\hline	
		\hline	
		\multirow{2}*{Variants}
		&\multicolumn{1}{c|}{HoSDR} &\multicolumn{1}{c|}{HoCDR} &\multicolumn{1}{c|}{HeSDR} &\multicolumn{1}{c}{HeCDR}\\
		\cline{2-5}	
		&P$\rightarrow$R  &A$\rightarrow$C  &\texttt{Ca} $\rightarrow$ \texttt{Am}	&\texttt{VisDA} \\ 
		\hline		
		$\mathcal{T}_1 + \mathcal{T}_2$ 
		&44.62  &17.11  &58.09  &9.67\\
		$\mathcal{T}_1 + \mathcal{T}_3$   
		&47.54 &18.65 &66.71   &13.69\\
		ATH\_U  
		&\textcolor{black}{51.70} 
		&\textcolor{black}{21.75}   &\textcolor{black}{75.64}    &\textcolor{black}{15.89}\\
		\hline
		\hline
	\end{tabular}
	\label{Table8}
\end{table}

\begin{figure}[!tb]
	\centering
	\subfigure[$\alpha_t$ vs. $\alpha_s$]{\includegraphics[scale=0.15,trim=360 35 370 110,clip]{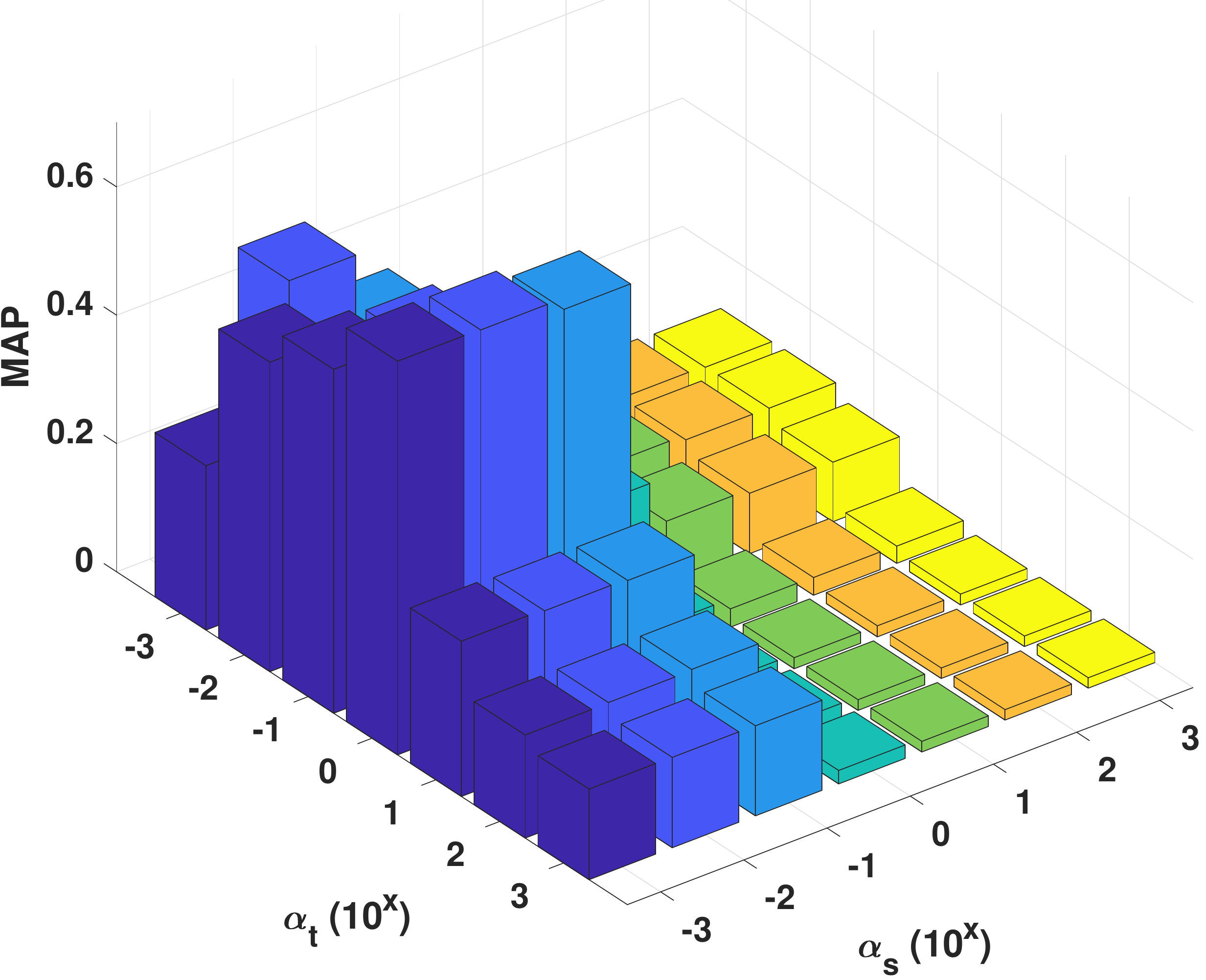}}
	\subfigure[$\beta_t$ vs. $\beta_s$]{\includegraphics[scale=0.15,trim=360 35 370 110,clip]{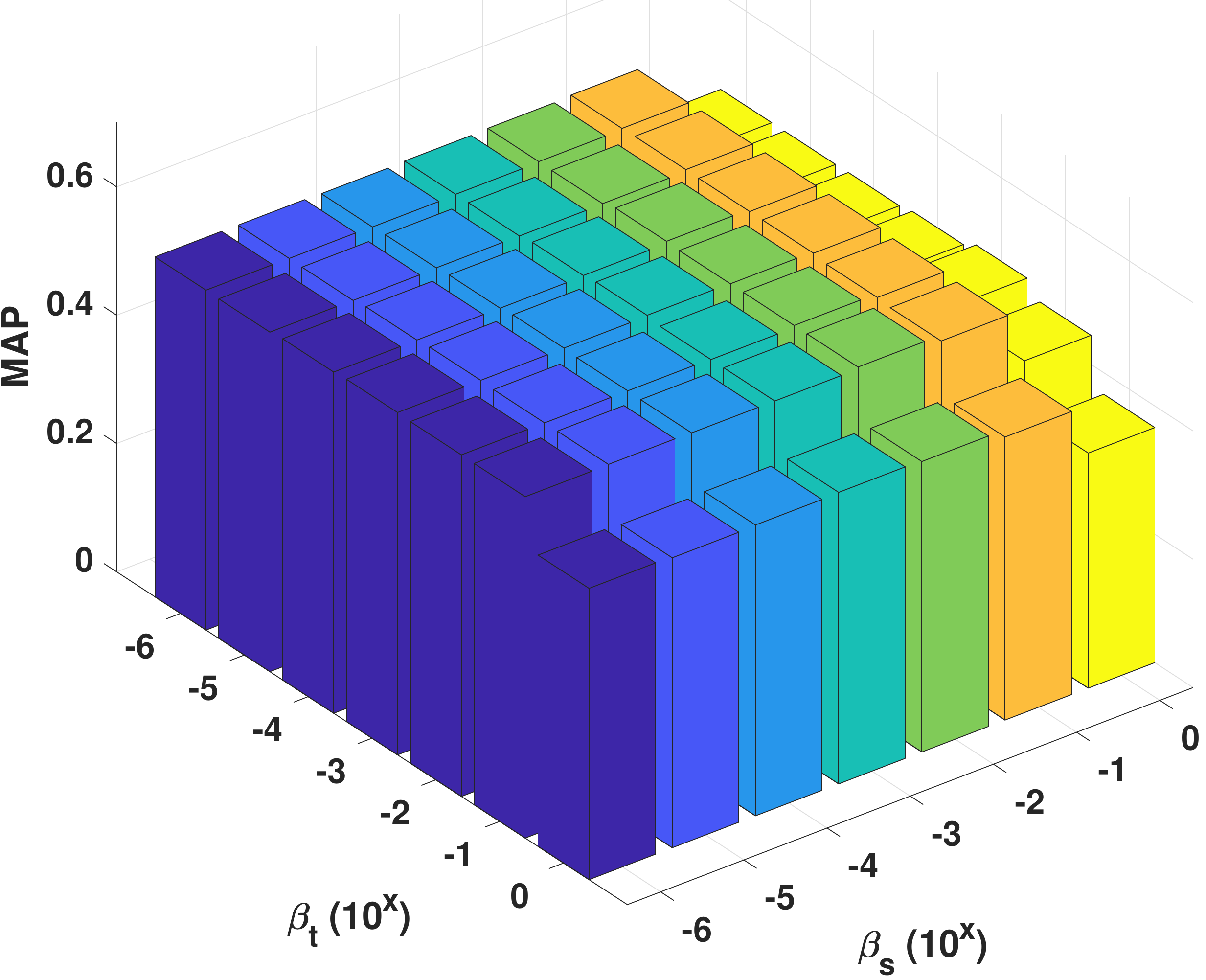}}
	\caption{Parameter sensitivity on A $\rightarrow$ C task.} 
	\label{parameters}
\end{figure}

\begin{figure*}[!tb]
	\centering
	\subfigure{\includegraphics[scale=0.34,trim=75 510 62 135,clip]{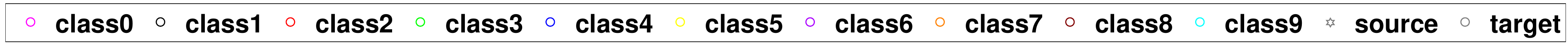}} \\
	\setcounter{subfigure}{0}
	\subfigure[GTH]{\includegraphics[scale=0.15,trim=80 55 70 50,clip]{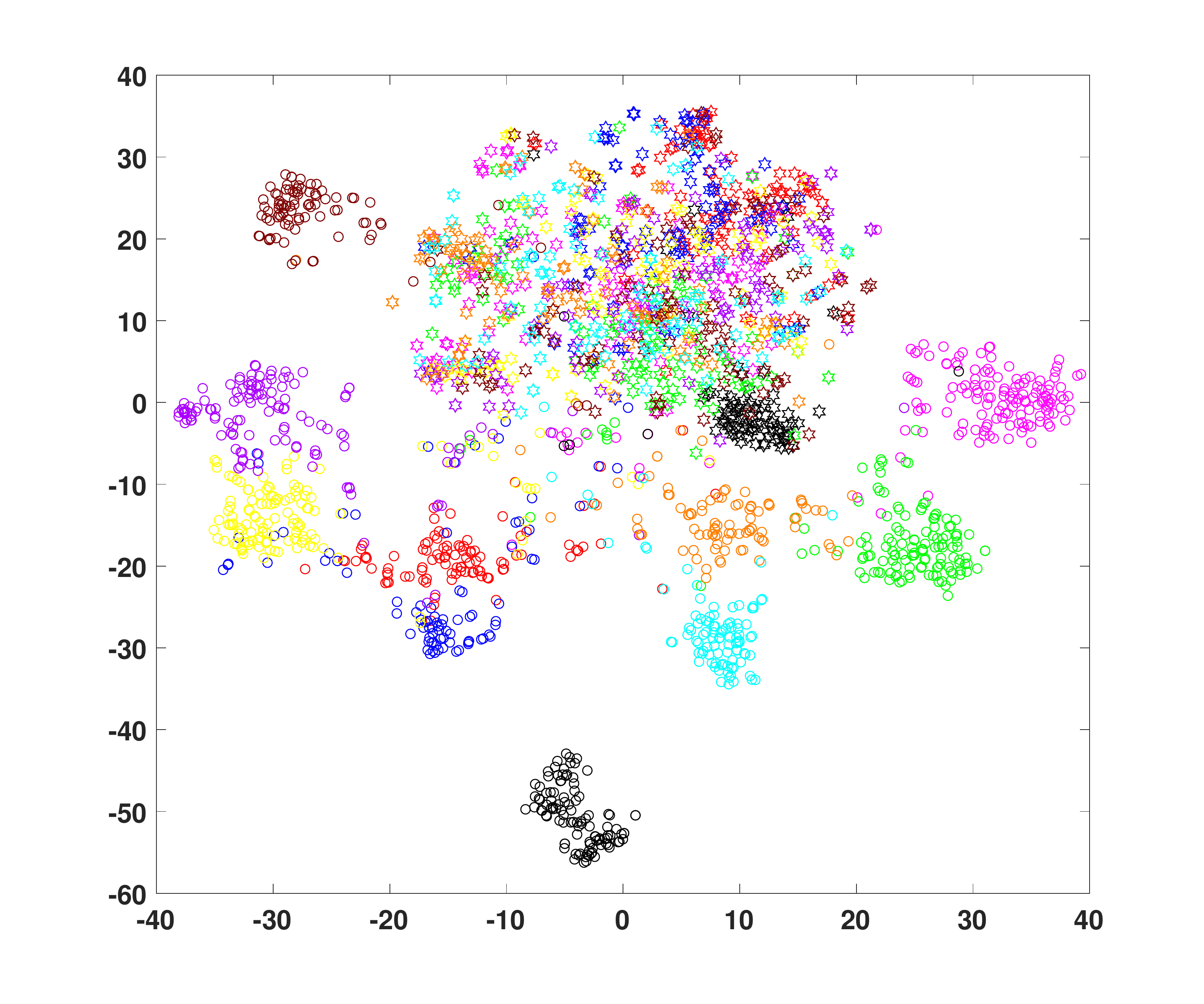}}
	\quad \subfigure[SDH]{\includegraphics[scale=0.15,trim=80 55 70 50,clip]{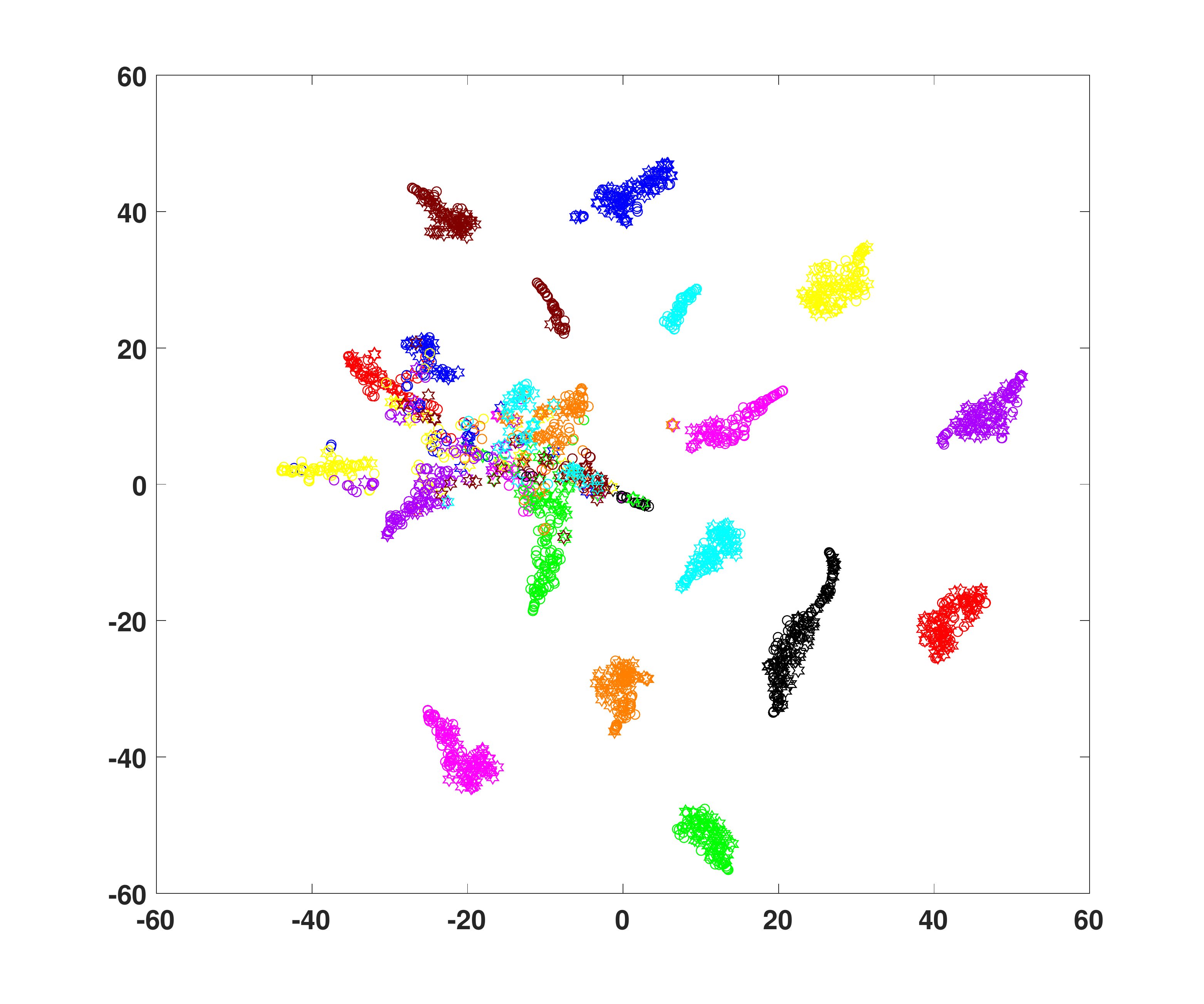}}
	\quad \subfigure[FSDH]{\includegraphics[scale=0.15,trim=80 55 70 50,clip]{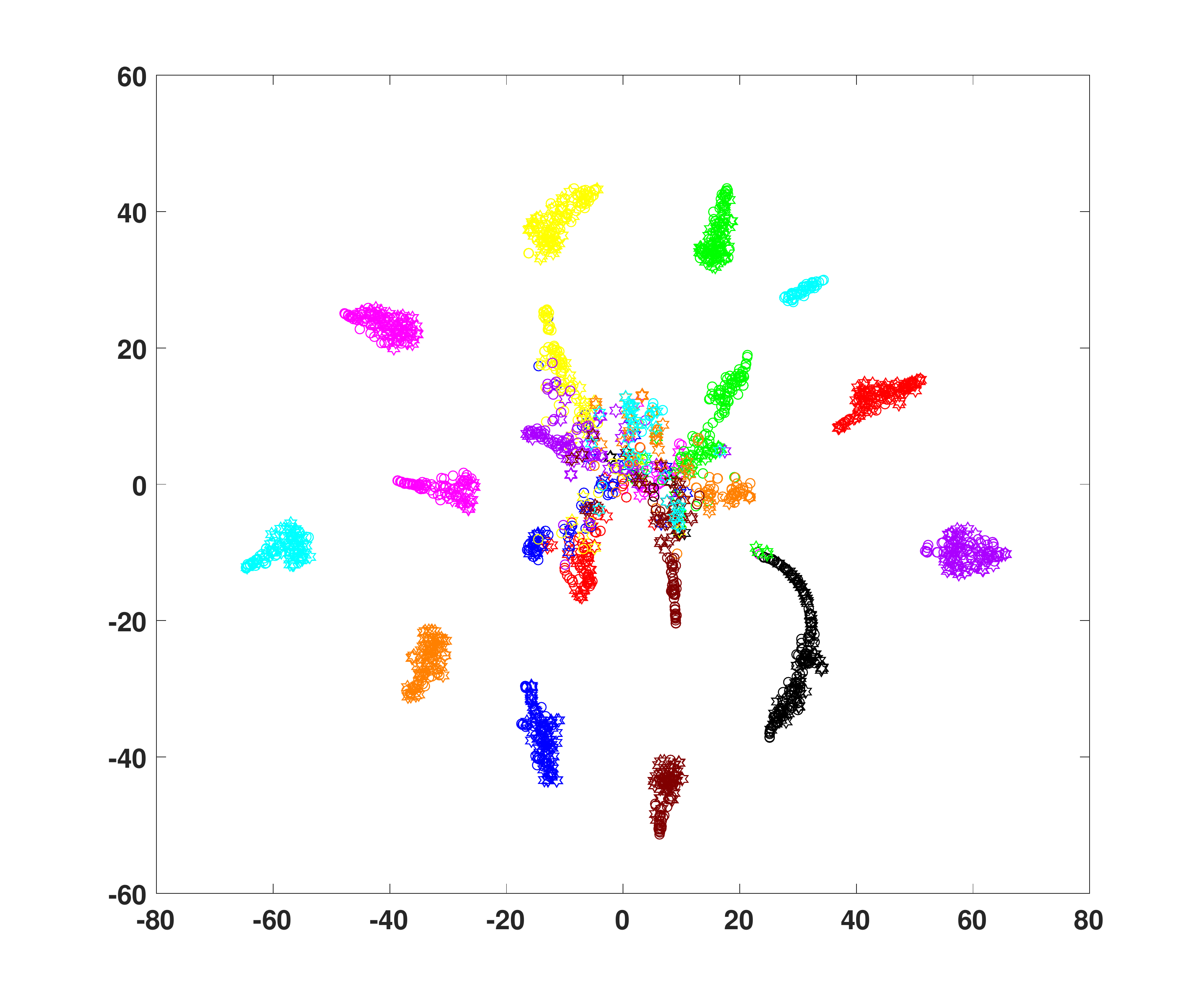}} \\
	\subfigure[ATH\_U]{\includegraphics[scale=0.15,trim=80 55 70 50,clip]{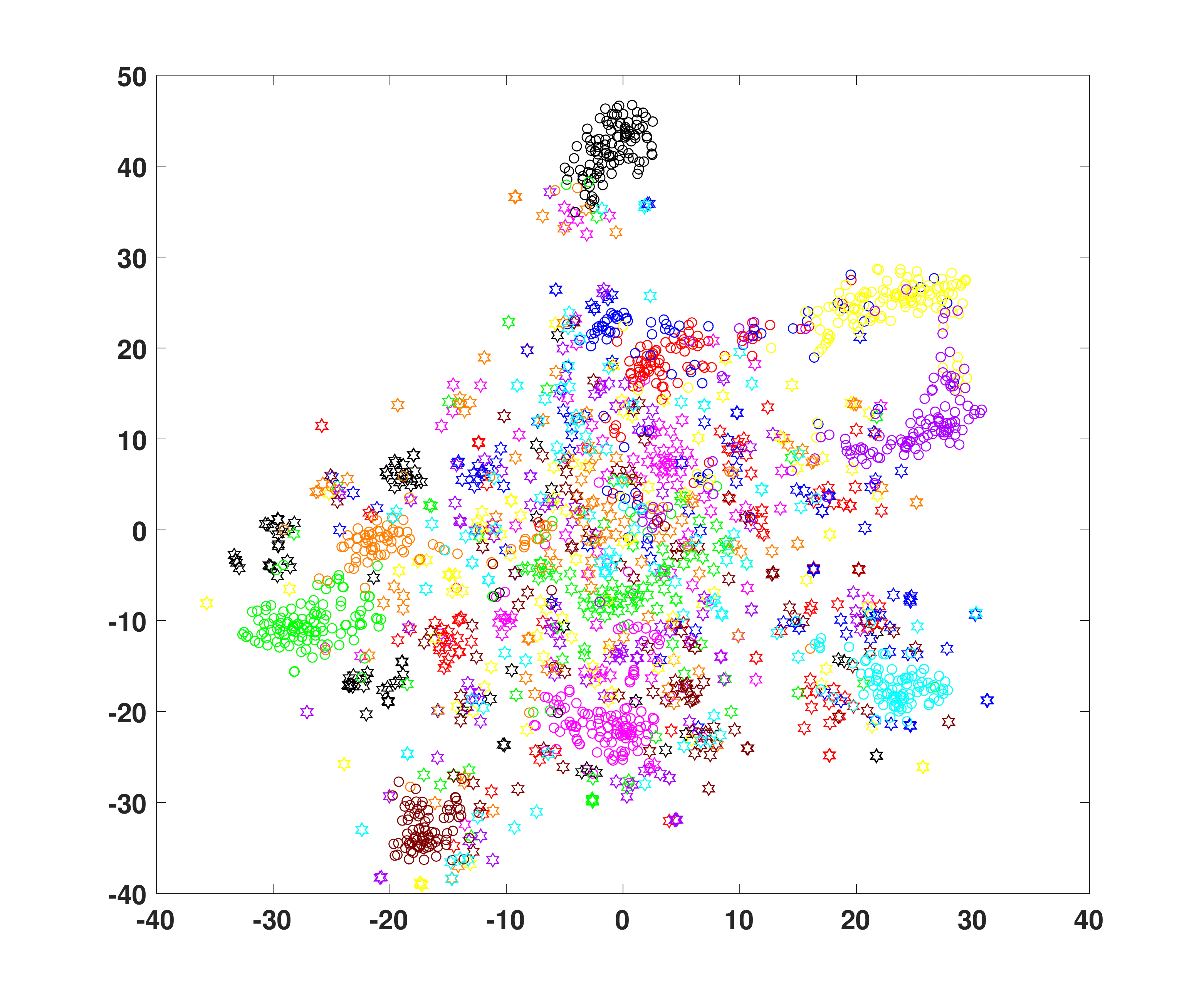}}
	\quad \subfigure[ATH\_K]{\includegraphics[scale=0.15,trim=80 55 70 50,clip]{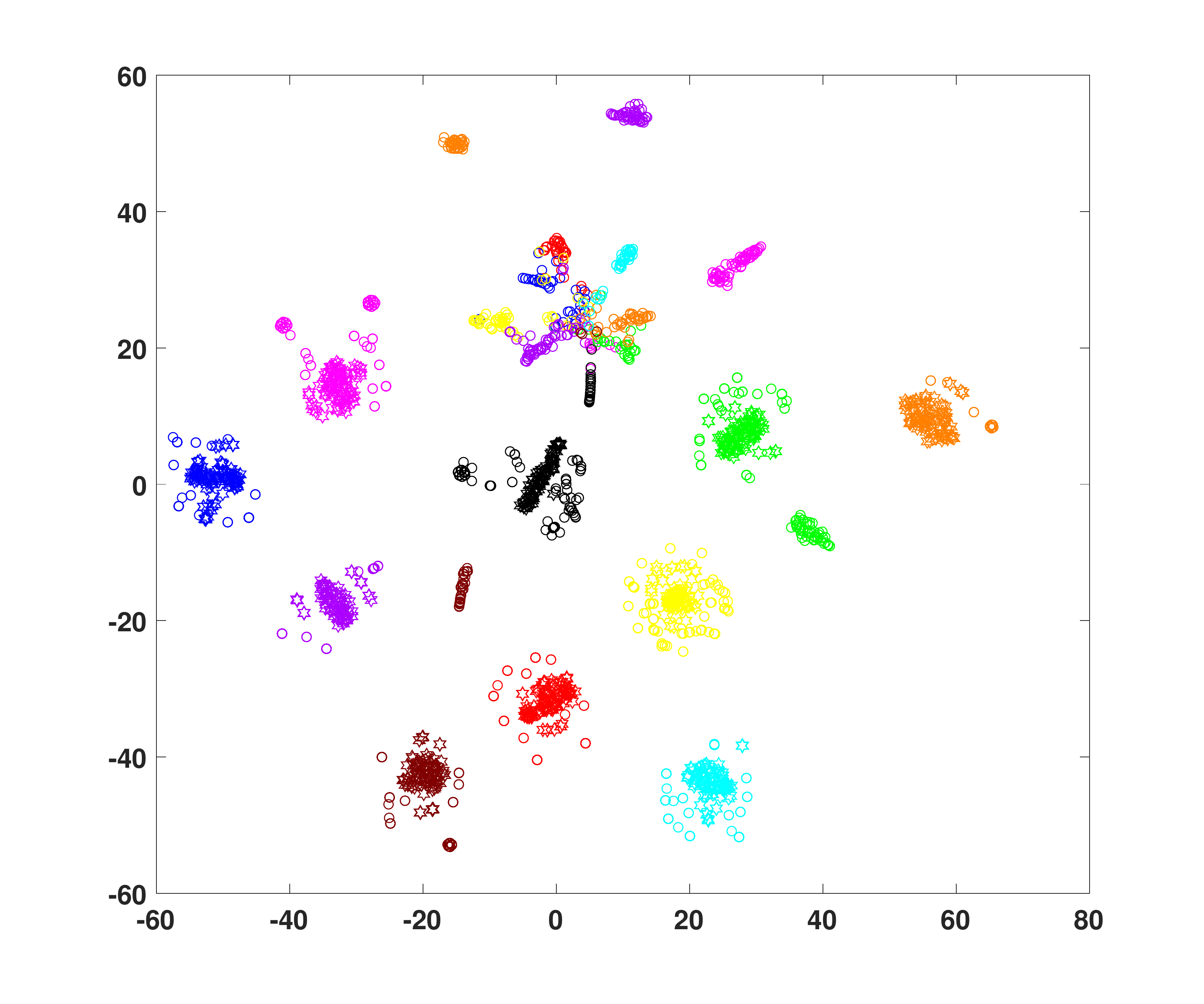}}
	\quad\subfigure[ATH\_S]{\includegraphics[scale=0.15,trim=80 55 70 50,clip]{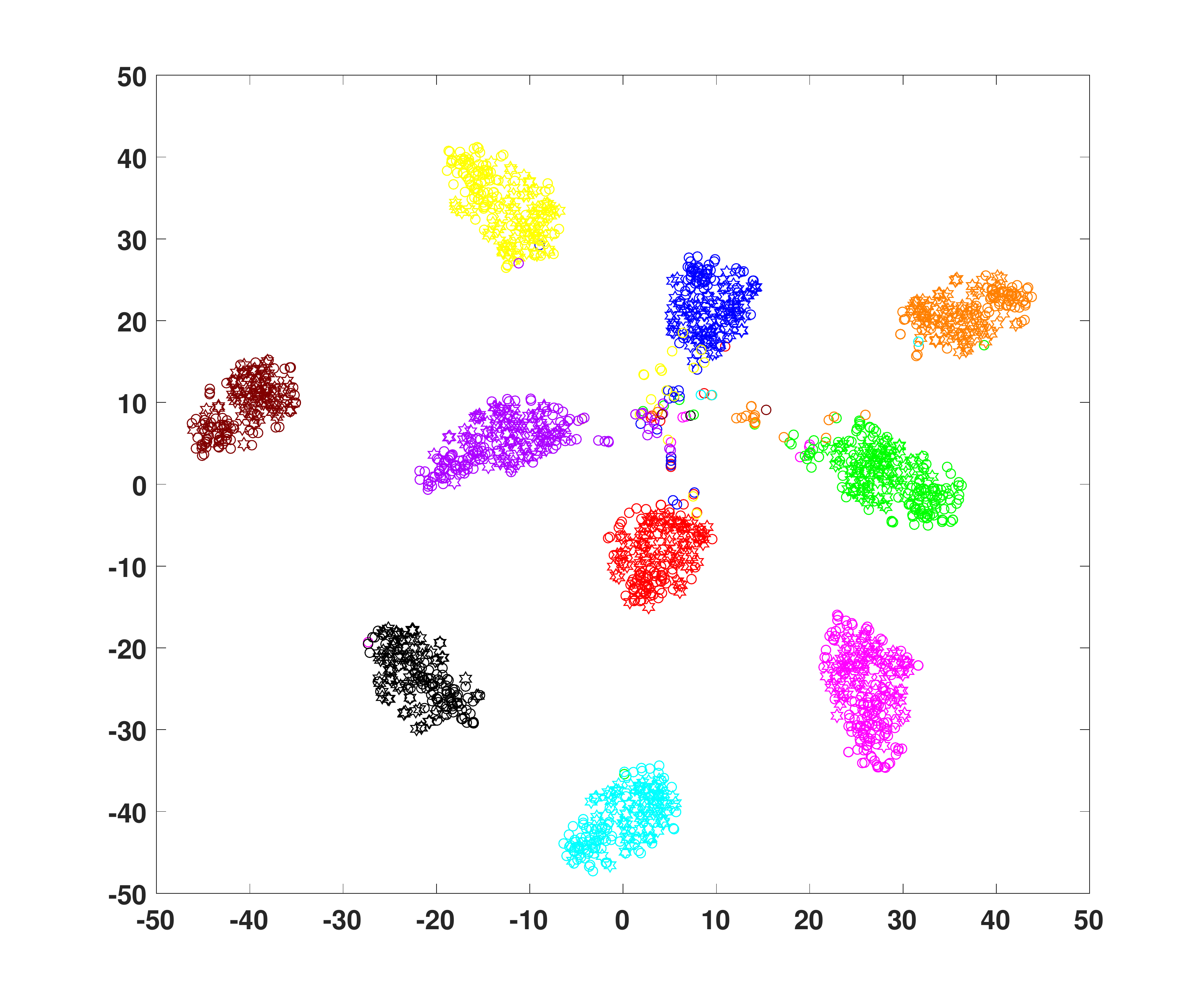}}
	\caption{Feature visualizations of different methods on the \texttt{Caltech} $\rightarrow$ \texttt{Amazon} task.}
	\label{tsne}
\end{figure*}

\subsection{Positive Transfer or Not}
We take ATH\_M as an example to study whether the proposed ATH framework can achieve positive transfer or not. 
Firstly, we use a $\mathtt{KNN}$ model ($k=1$) trained on labeled source domain to classify the unlabeled target domain data. In each iteration $ite$, the pseudo-labels $\widetilde{\mathbf{Y}}_t^{ite}$ of target domain samples are obtained by:
\begin{equation}
\widetilde{\mathbf{Y}}_t^{ite+1} = \mathtt{KNN}(\mathbf{A}_t^{T^{ite}}\mathbf{X}_t^{},   \mathbf{A}_s^{T^{ite}}\mathbf{X}_s^{}, \mathbf{Y}_s). 
\end{equation}
For initial $\widetilde{\mathbf{Y}}_t^{1}$, we set
\begin{small}
	\begin{equation}
	\widetilde{\mathbf{Y}}_t^{1}=\left\{
	\begin{array}{l}
	\mathtt{KNN}(\mathbf{X}_t, \mathbf{X}_s, \mathbf{Y}_s),\qquad\qquad\mathrm{if}\ \mathrm{homogeneous}, \\
	\mathtt{KNN}(\mathbf{A}_{t,pca}^{T}\mathbf{X}_t, \mathbf{A}_{s,pca}^{T}\mathbf{X}_s, \mathbf{Y}_s),\ \ \ \ \mathrm{otherwise.} 
	\end{array}
	\right.
	\end{equation}
\end{small}where $\mathbf{A}_{t,pca}$ and $\mathbf{A}_{s,pca}$ are initialized by PCA projections.
Then, in each iteration $ite$, the classification accuracy $Accuracy^{ite}$ with respect to pseudo-labels $\widetilde{\mathbf{Y}}_t^{ite}$ on target domain can be calculated by:
\begin{equation}
Accuracy^{ite} = \frac{\sum_{j}^{n_t} \mathcal{I}(\widetilde{\mathbf{Y}}_{t, :j}^{ite}, \mathbf{Y}_{t, :j}) }{n_t}, 
\end{equation}
where $\mathbf{Y}_{t, :j}$ is the $j$-th column of $\mathbf{Y}_{t}$, $\mathcal{I}(a, b)$ is an indicator function, if vector $a$ equals to $b$, $\mathcal{I}(a, b) = 1$, otherwise, $\mathcal{I}(a, b) = 0$.
Figure \ref{positive} shows the accuracy and convergence curves of the proposed ATH\_M on homogeneous (A $\rightarrow$ P) and heterogeneous (VisDA train $\rightarrow$ validation) tasks. As we can see, our designed optimization algorithm converges fast. Moreover, with the increase of iteration times, the accuracy gradually increases and finally reaches stability, which demonstrates the validity of the mapping space learned by the proposed ATH\_M. Therefore, our ATH\_M can achieve positive transfer during the training process, which is a potential reason that why our methods outperform the other comparison methods on different GITR subtasks.

\subsection{Parameter Sensitivity}
Figure \ref{parameters} shows the parameter sensitivity of the proposed ATH\_S on \texttt{Office Home} A $\rightarrow$ C task. In this experiment, we fix $\lambda = 10^{-2}$ and $r=16$. From Figure \ref{parameters}, we observe that the performance of ATH\_S is relatively stable to $\beta_k$ in the range of [$10^{-6}$, $10^{-1}$]. However, $\alpha_k$ affects the MAP scores of ATH\_S to a great extent. Specifically, ATH\_S can work well when $\alpha_s$ and $\alpha_t$ are in the ranges of [$10^{-3}$, $10^{-1}$] and [$10^{-2}$, $10^{0}$], respectively.

\subsection{Ablation Study}
Table \ref{Table8} lists ablation experimental results of the proposed ATH\_U. Obviously, ATH\_U obtains better performance under different GITR subtasks, which indicates the effectiveness of each component of ATH\_U. Moreover, without preserving domain structure  (i.e., $\mathcal{T}_3$), the MAP values of ATH\_U will decrease significantly in most cases, which demonstrates the necessity of maintaining the intrinsic geometrical structure of data in their respective domains.

\subsection{Visualization}
Figure \ref{tsne} shows t-SNE visualizations \cite{tSNE} learned by GTH, SDH, FSDH, ATH\_U, ATH\_K and ATH\_S on heterogeneous \texttt{Caltech} $\rightarrow$ \texttt{Amazon} task, where the hexagon and circle points indicate the projected data from source and target domains respectively. 
As we can see, without semantic information, GTH and ATH\_U cannot obtain distinct class structure. For SDH and FSDH, due to feature heterogeneity of data and information loss caused by dimensionality reduction, massive points of different classes are mixed together and hard to distinguish. More importantly, our ATH\_K and ATH\_S can make the source data of each class close to the target data of the same class, and far away from the data of different classes in both two domains. 

\begin{figure}[!tb]
	\centering
	{\includegraphics[scale=0.25, trim=0 0 0 0,clip]{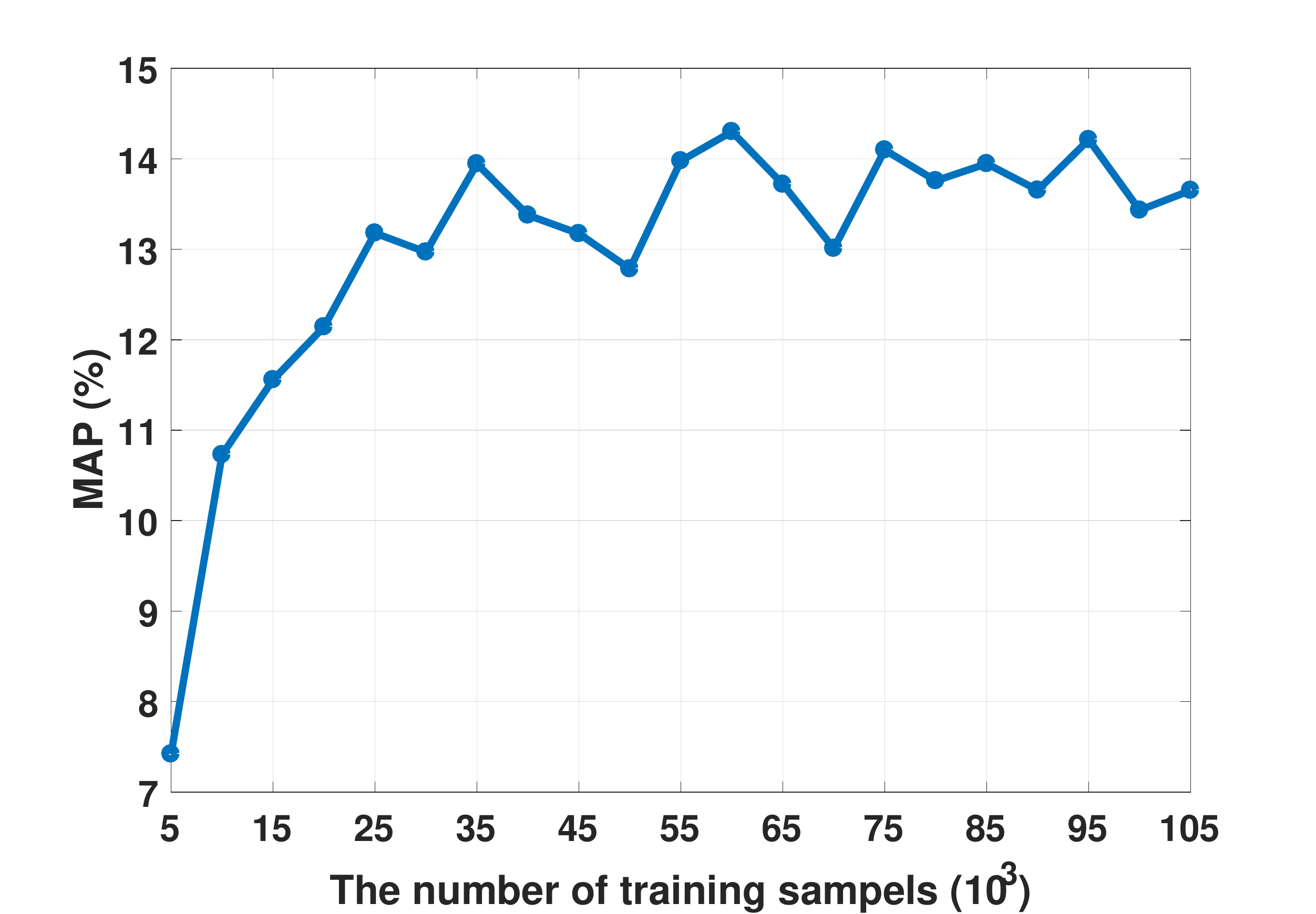}}
	\caption{The MAP results vs. the number of training sampels. } 
	\label{MAP}
\end{figure}

\subsection{Model Validation}
We have conducted additional experiments to show whether the validity of the proposed algorithm depends on the size of the training samples. Specifically, we conducted the experiments on the SUN397 dataset. In the experiments, we changed the number of training samples and calculated the MAP results on the test set. Figure \ref{MAP} shows the curve of MAP results ($16$ bits) changing with the number of training samples. It is easy to see that our algorithm can achieve good performance with only about $30\%$ of the data ($35\times 10^3$). Moreover, when the number of training samples increases to a certain extent, the performance of our method tends to be stable.

\section{Conclusion}
This paper presents a more difficult image transfer retrieval problem named GITR, which makes no restriction on the domains and feature spaces of query and retrieval sets. To solve the GITR problem, we propose an ATH framework with its unsupervised/semi-supervised/supervised realizations, based on asymmetric hashing strategy and adaptive bipartite graph learning. Extensive experiments conducted on different GITR subtasks and label settings demonstrate the effectiveness of our ATH framework. A potential limitation of the proposed ATH is that the class space of target domain needs to be aligned with that of source domain. For the class space misaligned problem, a possible solution is to integrate zero-shot learning approach into the proposed model, which is a potential research direction for our future work.

\bibliographystyle{IEEEtran}
\bibliography{IEEEabrv,ATH}

\end{document}